\documentclass[runningheads]{style-eccv/llncs}

\usepackage{style-eccv/eccv}

\usepackage{style-eccv/eccvabbrv}
\usepackage{graphicx}
\usepackage{booktabs}
\usepackage[accsupp]{axessibility}  

\usepackage{booktabs}   
\usepackage{tabularx}   
\usepackage{colortbl}   
\usepackage{multirow}
\usepackage{makecell}

\newcommand{\mc}[3]{\multicolumn{#1}{#2}{#3}}

\newcommand*{\rowstyle}[1]{
  \gdef\@rowstyle{#1}%
  \@rowstyle\ignorespaces%
}
\newcolumntype{=}{
  >{\gdef\@rowstyle{}}%
}
\newcolumntype{+}{
  >{\@rowstyle}%
}

\usepackage{wrapfig}
\usepackage{mwe} 
\usepackage{duckuments} 

\usepackage[font=normal,labelfont=bf]{caption}
\usepackage[font=normal]{subcaption}
\usepackage[leftcaption]{sidecap}  

\usepackage[normalem]{ulem} 
\usepackage{array}
\usepackage{xparse}
\usepackage{pifont}
\usepackage{ifthen}

\usepackage{textcomp}  
\usepackage{scalerel}  

\usepackage[pagebackref,breaklinks,colorlinks,citecolor=eccvblue]{hyperref}

\usepackage[capitalize]{cleveref}
\crefformat{section}{Section~#2#1#3}
\crefformat{subsection}{Section~#2#1#3}
\crefformat{figure}{Fig.~#2#1#3}
\crefformat{equation}{Eq.~#2#1#3}
\crefformat{table}{Table~#2#1#3}
\crefformat{algorithm}{Alg.~#2#1#3}
\usepackage{enumitem}

\DeclareEmphSequence{\bfseries\itshape}

\usepackage{orcidlink}
\newcommand{\method}{{\fontfamily{lmtt}\selectfont\emph{L-MIMO}}\xspace}

\newcommand{\std}[1]{\footnotesize $\pm{}$ #1}
\newcommand{\cmark}{\ding{51}}%
\newcommand{\xmark}{\ding{55}}%

\newlength\savewidth
\newlength\thinwidth

\definecolor{Gray}{gray}{0.93}
\definecolor{ForestGreen}{rgb}{0.13, 0.75, 0.13}

\usepackage{xspace} 
\makeatletter
\DeclareRobustCommand\onedot{\futurelet\@let@token\@onedot}
\def\@onedot{\ifx\@let@token.\else.\null\fi\xspace}
 
\def\ie{\textit{i.e}\onedot}

\makeatother

\usepackage{circledsteps}
\usepackage{amsmath,amsfonts,bm}

\def\1{\bm{1}}

\def\vm{{\bm{m}}}

\def\vp{{\bm{p}}}

\def\vz{{\bm{z}}}

\DeclareMathAlphabet{\mathsfit}{\encodingdefault}{\sfdefault}{m}{sl}
\SetMathAlphabet{\mathsfit}{bold}{\encodingdefault}{\sfdefault}{bx}{n}

\def\gL{{\mathcal{L}}}

\def\gR{{\mathcal{R}}}

\def\gT{{\mathcal{T}}}

\def\gV{{\mathcal{V}}}

\newcommand{\E}{\mathbb{E}}

\renewcommand{\paragraph}[1]{\noindent {\bf #1}}

\title{Towards Latent Masked Image Modeling for Self-Supervised Visual Representation Learning}

\author{%
Yibing Wei\inst{1} \and
Abhinav Gupta\inst{2} \and
Pedro Morgado\inst{1}
}

\authorrunning{Y.~Wei et al.}

\institute{
University of Wisconsin-Madison \and
Carnegie Mellon University
}

\begin{document}
\maketitle

\begin{abstract}
    Masked Image Modeling (MIM) has emerged as a promising method for deriving visual representations from unlabeled image data by predicting missing pixels from masked portions of images. It excels in region-aware learning and provides strong initializations for various tasks, but struggles to capture high-level semantics without further supervised fine-tuning, likely due to the low-level nature of its pixel reconstruction objective. 
    A promising yet unrealized framework is learning representations through masked reconstruction in latent space, combining the locality of MIM with the high-level targets. However, this approach poses significant training challenges as the reconstruction targets are learned in conjunction with the model, potentially leading to trivial or suboptimal solutions.
    Our study is among the first to thoroughly analyze and address the challenges of such framework, which we refer to as Latent MIM. Through a series of carefully designed experiments and extensive analysis, we identify the source of these challenges, including representation collapsing for joint online/target optimization, learning objectives, the high region correlation in latent space and decoding conditioning. By sequentially addressing these issues, we demonstrate that Latent MIM can indeed learn high-level representations while retaining the benefits of MIM models.
    Code is available at \url{https://github.com/yibingwei-1/LatentMIM}.
\end{abstract}


\section{Introduction\label{sec:intro}}

Masked Image Modeling (MIM), a learning framework that derives visual representations from unlabeled image data, has recently gained prominence. This technique masks a substantial part of an image and trains a model to predict the missing pixels using the surrounding context. Despite the simple learning objective, MIM has been shown to learn powerful representations, which, when fine-tuned, can achieve state-of-the-art performance on a variety of downstream tasks, including object classification, detection, and segmentation~\cite{he2022masked,bao2021beit}.
MIM approaches offer several key advantages over other self-supervised visual representation learning methods. By requiring the model to accurately reconstruct all masked patches, MIM incentivizes the model to maintain distinct local representations of each image region and forces the model to reason over the spatial layout of object subparts. 
These benefits make MIM a popular approach for learning visual representations in a self-supervised manner.

However, the representations learned through this framework fail to capture high-level semantics without further supervised fine-tuning, as shown by their lower performance in linear probing and nearest-neighbor classification. We hypothesize that MIM does not directly learn high-level semantics due to the use of low-level learning targets like raw pixels. To effectively reconstruct the high-frequency details, the learned representations must retain a low-level description of the image, which limits their ability to encode higher-level semantics.

\begin{figure}[t!]
    \centering
    \includegraphics[width=\linewidth]{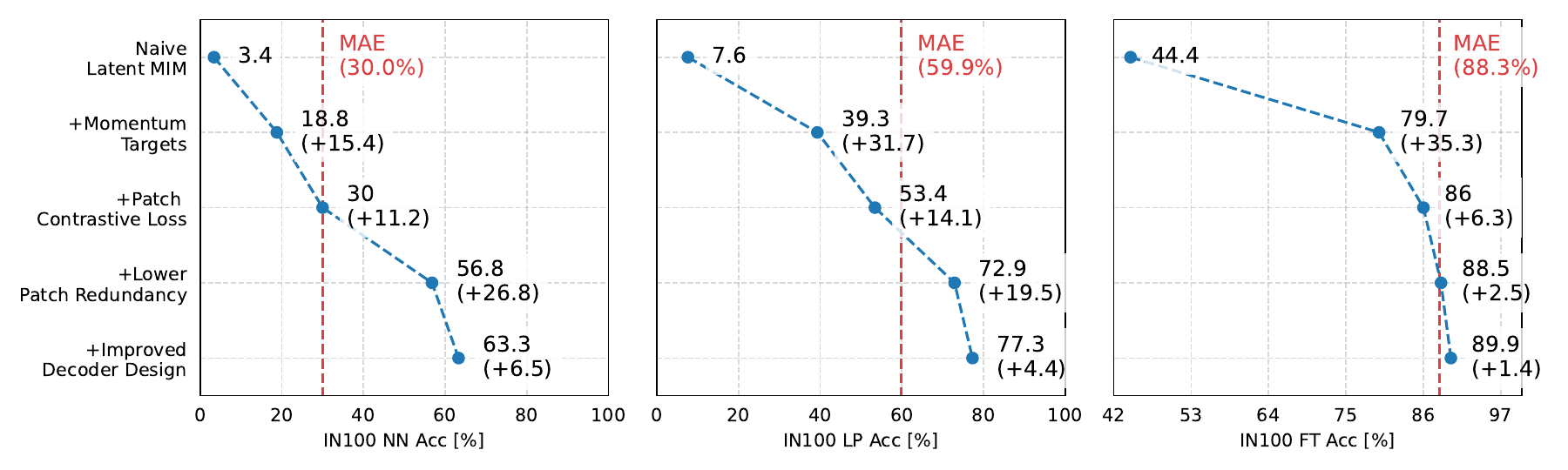}
    \caption{\textbf{Challenges of Latent MIM.} The representations learned by MIM approaches fail to capture high-level semantics, as shown by the poor performance in nearest neighbor and linear probe evaluation.\label{fig:study-overview}}
    
\end{figure}
%



A natural solution to this challenge is to perform \emph{masked image modeling in latent space}, a general framework that we refer to as Latent MIM.~This means that instead of reconstructing raw pixels, Latent MIM methods should learn representations by defining local (patch-wise) latent representations for each image and reconstructing the representations of masked regions from visible ones. By bypassing low-level pixel targets and focusing on distinguishing the latent representations of masked regions, Latent MIM methods hold the promise of substantially enhancing the semantics obtained through pixel-level MIM approaches and potentially their downstream performance.

However, intuitive adaptations of MIM to latent space reconstruction are unstable due to (1) the \emph{existence of trivial solutions} and (2) the \emph{high correlation between the semantics of nearby patches}. First, trivial solutions occur when the learned latent devolves into easy-to-predict but uninformative representations of the input images; for example, using a mean squared error (MSE) loss, the decoder might generate constant latent representations across all regions. Second, even if latent representations are indicative of the contents of each region, the semantics of nearby patches are more correlated with each other than the pixels themselves. This can also lead to poor representations, as the model can learn to predict masked regions by simply copying from nearby visible regions. As a result, despite its great potential, these challenges have significantly impeded the advancement of Latent MIM

In this work, we conduct an in-depth investigation into the challenges of pure latent MIM.~Our objective is to provide a thorough analysis, identifying, characterizing, and exploring potential solutions to the core challenges. 
\cref{fig:study-overview} provides an overview of our analysis, showing the downstream classification performance as we progress through four challenges, with significant improvements in nearest neighbor, linear probe, and fine-tuning accuracy.

\begin{description}[leftmargin=1.5em]
    \item[\Circled{1}] {\it Joint optimization of visible and masked regions representations leads to representation collapse.} Instead, as popularized in contrastive learning methods, creating asymmetries between the two representations, while avoiding the target encoder to contribute to the gradient computation, is crucial for learning meaningful representations.
    \item[\Circled{2}] {\it Direct reconstruction results in poor representations, regardless of the specific loss used}. Beyond direct reconstruction, we explore contrastive predictive coding within image patches\footnote{We explicitly avoid batch-wise contrastive objectives to focus our study on pure masked reconstruction.}, which not only encourages the model to predict representations that are similar to those of target patches (thus achieving the reconstruction objective) but also encourages richer and spatially diverse representations across the image.
    \item[\Circled{3}] {\it Controlling the redundancy between visible and target patches is critical.} Specifically, the optimal masking strategies for MIM in pixel space (\ie, 75\% masking of regular grids) can result in a set of target patches whose representations are too similar to (and thus easily predicted from) visible patches. To reduce the redundancy between patches, we investigate different procedures to generate the visible and target token sequences, including the use of non-contiguous grids, where nearby patches are separated by a random number of pixels, higher mask ratios and explicit patch similarity constraints.
    \item[\Circled{4}] {\it The decoder design must be carefully crafted since Latent MIM decoders predict high-level representations of mask regions from those of visible regions.} The decoder needs to have enough capacity to fill in the missing information but is simultaneously shallow enough to avoid taking on the role of the encoder (\ie, computing the image semantics). To this end, we investigate a variety of decoder architectures. We observe that cross-attention decoders with modified mask tokens that directly encode the representation of nearby visible regions were particularly effective.
\end{description}

By explicitly investigating the inherent challenges of Latent MIM, we demonstrate that the Latent MIM framework can indeed be used to learn richer semantics and diverse patch-wise representations compared to existing MIM approaches, without the need for supervised fine-tuning. We further validate our findings by scaling up an instantiation of the Latent MIM informed by our findings to ImageNet-1K.
The learned representations achieved 50.1\% nearest neighbor (a 37.9\% gain over MAE) and 72.0\% linear probing (+4.2\% gain over MAE) top-1 classification accuracy.
The semantics and localizability of the learned representations are further demonstrated in three distinct tasks that require robust local representations: unsupervised scene segmentation, video object segmentation, and few-shot transfer learning across a variety of tasks. Remarkably, representations learned through Latent MIM approaches are capable of generating accurate segmentation of visual scenes, even in the absence of supervision.
In video object segmentation and few-shot transfer learning, Latent MIM surpasses both the pixel-level MIM techniques, such as MAE, and previous Latent MIM methodologies like data2vec.

\section{\label{sec:latent-mim}Latent Masked Image Modeling}


\subsection{Framework Overview} 
\begin{figure}[t!]
    \centering
    \includegraphics[width=0.9\linewidth]{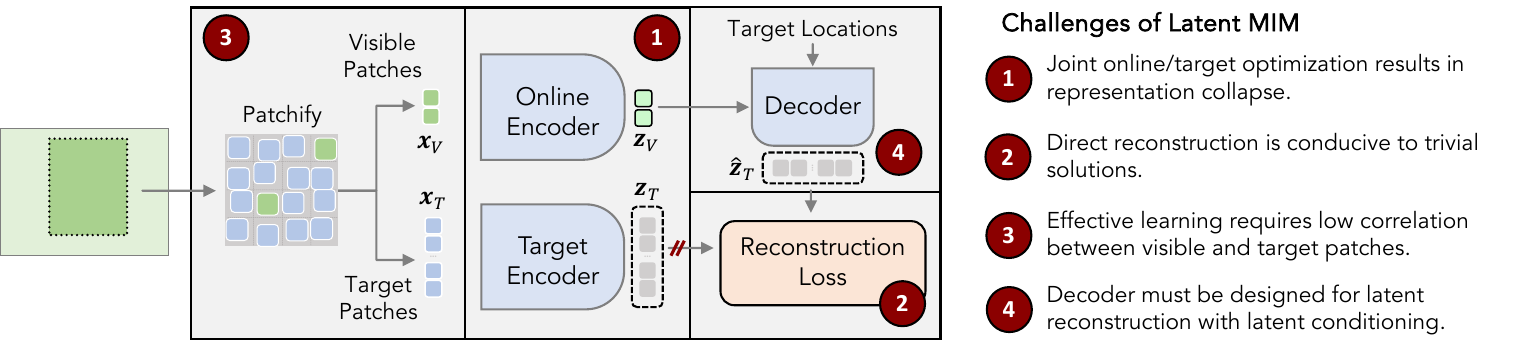}
    \caption{\label{fig:overview-method}\textbf{Latent Masked Image Modeling Overview}. Models are trained to reconstruct the latent representations generated by a target encoder at withheld locations. Four major challenges for effectively deploying Latent MIM are identified in this work, as well as potential solutions. These challenges relate to joint encoder optimization, direct reconstruction loss, the semantic correlation between visible and target patches, and the decoder design.}
    
\end{figure}

Latent MIM is a self-supervised learning framework that aims to learn visual representations through masked image modeling in latent space. 
As illustrated in \cref{fig:overview-method}, Latent MIM models comprise three components: an \textit{online encoder} $f(\cdot)$, a \textit{target encoder} $f_T(\cdot)$, and a \textit{decoder} $g(\cdot)$. 
Similar to pixel-based MIM approaches~\cite{he2022masked,bao2021beit}, each image is first divided into a set of $L$ small patches $x_i$ along with their location within the image $p_i$. This set $X=\lbrace (x_i, p_i)\rbrace_{i=1}^L$ is randomly split into two disjoint sets: the \textit{visible} $X_\gV=\lbrace (x_i, p_i)\rbrace_{i\in\gV}$ and \textit{target} $X_\gT=\lbrace (x_i, p_i)\rbrace_{i\in\gT}$ patches, where $\gV$ and $\gT$ are non-overlapping index sets of grid locations. As the name suggests, the visible patches are used to generate the latent representation of an image, while the target patches are used as the reconstruction targets.
To accomplish this, the online encoder extracts latent representations of the visible patches 
\begin{equation}
    \lbrace \vz_i\rbrace_{i\in\gV} = Z_\gV = f(X_\gV),
    \label{eq:online-encoder}
\end{equation}
which are then used to inform the decoder for predicting patches at the target locations $P_\gT=\lbrace p_i \rbrace_{i\in\gT}$
\begin{equation}
    \hat{Z}_\gT = g(Z_\gV, P_\gT).
    \label{eq:decoder}
\end{equation}
However, instead of predicting pixel values, Latent MIM imposes the latent representations obtained from the \textit{target encoder} as the reconstruction targets,
\begin{equation}
    \lbrace \vz_i\rbrace_{i\in\gT} = Z_\gT = f_T(X_\gT).
    \label{eq:target-encoder}
\end{equation} 
Latent MIM models can then be trained to minimize the discrepancy $\Delta$ between the predicted target representations $\hat{Z}_\gT$ and those obtained from the target encoder $Z_\gT$. This is achieved by a reconstruction loss
\begin{equation}
    \gL_{rec}=\E\left[\Delta\left(Z_\gT, \hat{Z}_\gT\right)\right],
    \label{eq:reconstruction-loss}
\end{equation}
where the expectation is taken over the training dataset.
In pixel-based MIM, the mean squared error (MSE) is a popular choice for $\Delta$, $\Delta = \frac{1}{|\gT|}\sum_{i\in\gT}\|\hat{\vz}_i - \vz_i\|^2$. However, as shown in \cref{sec:challenge2}, the MSE loss is not effective for Latent MIM.

\subsection{Distinctions and similarities from related frameworks}
\paragraph{Masked Image Modeling (MIM)}  methods learn by predicting masked parts of an image using targets derived from simple transformations of the original image.  
Targets can be raw pixels~\cite{he2022masked, zhang2022point}, hand-crafted features such as HOG~\cite{wei2022masked} or pre-trained features ~\cite{bao2021beit}. Thus, low-level MIM is required to maintain localized representations of an image. However, the model capacity is partially consumed by low-level details, limiting its capacity for high-level semantics.

\paragraph{Contrastive Learning (CL)}
In contrast, CL learns representations of each image in the context of the dataset in which they occur~\cite{mocov3,dino,byol}. While CL can achieve outstanding semantics without finetuning, learning in context can be undesirable, for example, when the data distribution does not provide appropriate negative samples that highlight meaningful semantic distinctions between images. Batch dependencies also make contrastive objectives less flexible and reproducible, as they depend on the available compute resources. Instead, pure Latent MIM objectives learn representations from the image itself, and independently from other images. 

\paragraph{Previous explorations for Latent MIM}
The potential of Latent MIM is tied to its ability to learn from high-level and region-aware targets, which are continuously improved throughout training.  However, prior works attempting to deploy Latent MIM e~\cite{dong2023peco,zhou2021ibot,tao2023siamese,chen2023context,baevski2022data2vec,yi2022masked} do not directly tackle the core optimization challenges inherent in Latent MIM discussed in \cref{sec:challenges}. These challenges have been mitigated by treating latent MIM as a supplementary objective to other techniques, such as global contrastive learning~\cite{zhou2021ibot,tao2023siamese}, low-level reconstruction~\cite{chen2023context}, or alternatively, by using fixed pre-trained features as the latent targets~\cite{bao2021beit,dong2023peco}, instead of jointly learning the target and online representations. 
 
The most closely related prior works focused on pure latent MIM, include data2vec~\cite{baevski2022data2vec} ConMIM~\cite{yi2022masked} and I-JEPA~\cite{assran2023self}. Despite their contribution, data2vec and ConMIM use mask tokens instead of patch removal to hide image regions, causing a mismatch between pre-training and real-world deployment with unmasked images, which leads to poor performance in downstream tasks without finetuning. I-JEPA learns by predicting contiguous regions in latent space via an MSE loss, potentially leading to low-resolution (``blurred'') or less localizable semantics (one of the major benefits of pixel-based MIM models).

While these works have advanced beyond pixel-level MIM, they have not thoroughly analyzed or addressed the inherent challenges specific to Latent MIM. Therefore, we argue that the full potential of this framework remains untapped, positioning Latent MIM as a significant, high-reward research area. Latent MIM has the potential to generate rich and localizable high-level semantic representations while maintaining the diverse patch-wise representations characteristic of MIM approaches.

\section{\label{sec:challenges}Challenges of Latent MIM}
To better focus on the various difficulties of Latent MIM optimization, we introduce and analyze each challenge separately. We begin by describing a naive implementation that fails to learn meaningful representations. We then introduce each of the four challenges (\cref{sec:challenge1}-\ref{sec:challenge4}) and conduct a thorough analysis of potential strategies to address them. Each section builds on the findings of the preceding one, yielding increasingly effective instantiations of the Latent MIM framework. \cref{fig:study-overview} shows the overview of performance through this progression.

\subsection{\label{sec:study-design}Experimental Design}
\paragraph{Pre-training} For a comprehensive experimental analysis within an acceptable compute budget, we conduct experiments using the standard ViT-B transformer backbone for both the online and target decoders and trained the model on the ImageNet-100 (IN100) dataset, a subset of ImageNet-1k, containing 100 classes selected at random. The class partition used follows that of~\cite{looc,cmc}. This dataset contains approximately 125,000 images, sufficient for executing experiments with statistical significance. Section~\ref{sec:in1k} further discusses the scaling properties of the proposed framework in the context of larger datasets, like ImageNet-1k.

All models were pre-trained for 300 epochs using the AdamW~\cite{adamw} optimizer with a batch size of 1024, a base learning rate of $1.5 \times 10^{-4}$, following 30 warm-up epochs and a cosine decay schedule. Data augmentations were applied to the input images, including random horizontal flipping and random image cropping with a minimum crop area of 0.2.

\paragraph{Downstream tasks} We evaluated all models using nearest-neighbor, linear probing, and fine-tuning on the same dataset. For \emph{nearest-neighbor} evaluation, we extract the representations for all images in the test set and report the fraction of samples whose nearest neighbor shares the same class.
For \emph{linear probing}, we train for 20 epochs using the LARS optimizer~\cite{you2017large} with a base learning rate of 0.5, 2 warm-up epochs, and a batch size of 1024. 
For \emph{fine-tuning}, we train for 50 epochs using an AdamW~\cite{adamw} optimizer with a base learning rate of 0.001, a weight decay of 0.05, 5 warm-up epochs, and a batch size of 1024.

\subsection{Naive Latent MIM}
To fully appreciate the complexities involved in learning meaningful representations through Latent MIM, we initiate our discussion with a simple and intuitive implementation, by closely following those of pixel-based MIM models~\cite{he2022masked}.
Input images of resolution 224$\times$224 are divided into a regular 14$\times$14 grid of patches and partitioned into a 25/75\% split to create the sets of visible/target patches, respectively.
We then use a standard ViT-B transformer~\cite{vits} to encode both the visible and target patches ($f$ and $f_T$) and a transformer decoder $g$ to reconstruct the target patches from the visible latent representations. We use the same self-attention decoder as in MAE~\cite{he2022masked}, but reduce its depth to only 3 layers since, in Latent MIM, the decoder task (predicting latents from latents) is less complex than in pixel-based MIM (predicting pixels from latents).
Both online and target encoders and the decoder are jointly trained using the mean squared error (MSE) loss between the predicted and target latents.

\begin{figure}[t]
    \centering
    \begin{minipage}{0.7\linewidth}
        \centering
        \caption{{\bf Training Collapse of the Naive Latent MIM.} This solution achieves a zero reconstruction loss but fails to capture any meaningful information about the input images. As a result, the nearest neighbor (NN) evaluation yields random performance. Top: NN accuracy; Bottom: training loss.\label{fig:challenge1}}
    \end{minipage}\hfill
    \begin{minipage}{0.27\linewidth}
        \centering
        \includegraphics[width=\textwidth]{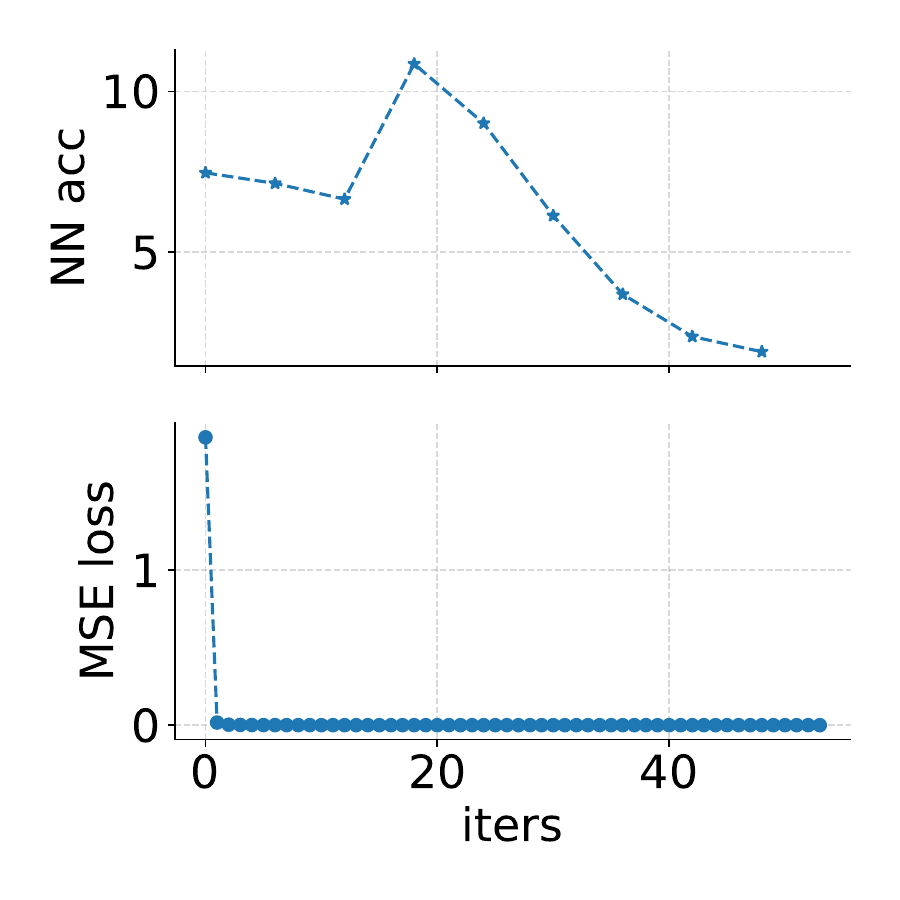} 
    \end{minipage}
\end{figure}

As shown in \cref{fig:challenge1}, the Naive Latent MIM model is highly unstable. We observe that within a short number of iterations, the model collapses into a degenerate solution. The subsequent sections seek to understand the underlying causes of this finding and propose strategies to mitigate it.



\newcommand{\loss}{
    \begin{tabular}{cccc}
        \toprule
        
        \bf Loss  & \bf NN & \bf LP & \bf Ft \\
        \midrule
        MSE & 18.8 & 39.3 & 79.7 \\
        L1 & 15.2 & 36.7 & 81.7 \\
        Huber & 24.3 &46.4 &  82.1  \\
        PatchDisc & \bf 30.0 & 53.4 & 86.0 \\
        \midrule
        MAE~\cite{he2022masked} &  \bf 30.0 &  \bf 59.9 &  \bf 88.3 \\
        \bottomrule
    \end{tabular}
}


\newcommand{\gap}{
\begin{tabular}{cccc}
    \toprule
    \bf G & \bf NN & \bf LP & \bf Ft \\
    \midrule
    0 &  53.3 & 70.4 & 87.6\\
    2 &  56.2 & 72.5 & 88.4 \\
    4 & \bf 56.8  & \bf 72.9 & \bf 88.5 \\
    6 &  53.4 & 70.4 & 87.3 \\
    \bottomrule
\end{tabular}
}

\newcommand{\maskratio}{
    \begin{tabular}{cccc}
        \toprule
        \bf Mask & 
        \bf NN & \bf LP & \bf Ft \\
        \midrule
        0.5 & 5.6 & 9.5 &  54.2\\
        0.75 & 30 & 53.4& 86.0 \\
        0.9 & \bf 53.3& \bf 70.4 & \bf 87.6 \\
        0.95 & 38.8 & 59.5 &  83.3 \\
        \bottomrule
    \end{tabular}
}

\newcommand{\sparsetgt}{
    \begin{tabular}{cccc}
        \toprule
        \bf K & 
        \bf NN & \bf LP & \bf Ft \\
        \midrule
        176 & 54.6 & 71.2 & 88.0 \\
        132 & \bf54.6 & \bf74.5 & \bf89.6 \\
        88 &  49.4 & 66.7& 86.2 \\
        44 & 27.4 & 54.2 & 81.1 \\
        \bottomrule
    \end{tabular}
}

\newcommand{\avgsimconstrain}{
    \begin{tabular}{cccc}
        \toprule
        \bf $\gamma$ & 
        \bf NN & \bf LP & \bf Ft \\
        \midrule
        none & 59.9 & 74.0 & 88.7  \\ 
        0.25$\rightarrow$0.75 & \bf 56.8 & \bf 72.9 & \bf 88.5 \\ 
        \bottomrule
    \end{tabular}
}

\newcommand{\combinedcorrelation}{
    \begin{tabular}{cccccc}
        \toprule
        \bf Mask &\bf Gap &\bf Sim. & 
        \bf NN & \bf LP & \bf Ft \\
        \midrule
        0.75 & 0 & \xmark{} & 30.0 & 53.4 & 86.0 \\
        0.9  & 0 & \xmark{} & 53.3 & 70.4 & 87.6 \\
        0.9  & 4 & \xmark{} & 54.6 & 71.2 & 88.0 \\
        0.9  & 4 & \cmark{} & \bf 56.8 & \bf  72.9 &  \bf 88.5 \\
        \midrule
        \mc{3}{c}{MAE~\cite{he2022masked}} & 30.0 & 59.9 & 88.3 \\
        \bottomrule
    \end{tabular}
}


\newcommand{\projector}{
    \begin{tabular}{cccc}
        \toprule
        \bf Projector & 
        \bf NN & \bf LP & \bf Ft \\
        \midrule
        None & 56.88 & 72.8 & 88.5 \\
        MLP & \bf 59.7 & \bf 76.8 & \bf 89.4  \\
        \bottomrule
    \end{tabular}
}

\newcommand{\conditioning}{
    \begin{tabular}{ccccc}
        \toprule
        \bf Type & \bf VisCues &
        \bf NN & \bf LP & \bf Ft \\
        \midrule
        Self & \xmark{} & 59.7 & 76.8 & 89.4 \\
        Cross & \xmark{} & 61.1 & 77.0 & 89.4 \\
        \midrule
        Cross & \cmark{} &\bf 63.3 & \bf 77.3 & \bf 89.9  \\
        \bottomrule
    \end{tabular}
}

\newcommand{\decdepth}{
    \begin{tabular}{cccc}
        \toprule
        \bf Depth & 
        \bf NN & \bf LP & \bf Ft \\
        \midrule
        2 & 50.6 & \bf 77.8 & \bf 89.9 \\
        3 & \bf 63.3 &  77.3 & \bf 89.9  \\
        8 & 35.9 & 65.8 & 86.2 \\
        \bottomrule
    \end{tabular}
}

\newcommand{\combineddecoder}{
    \begin{tabular}{cccccc}
        \toprule
        \bf Decoder &  \bf Proj. & Depth & 
        \bf NN & \bf LP & \bf Ft \\
        \midrule
        Self-attn & none & 3 &  56.8 & 72.9 & 88.5 \\
        Self-attn & mlp & 3 & 59.7 & 76.8 & 89.4 \\
        Cross-attn &  mlp & 3 & 61.1 & 77.0 & 89.4 \\
        Cross-attn w/ Vis Cues &  mlp & 3 &   \bf 63.3 &  \bf 77.3 &  \bf 89.9 \\
        Cross-attn w/ Vis Cues &  mlp & 8 &  35.9 &  65.8 & 86.2 \\
        \midrule
        MAE~\cite{he2022masked} &  & 8 & 30.0 & 59.9 & 88.3 \\
        \bottomrule
    \end{tabular}
}


\newcommand{\tolmimo}{
    \begin{tabular}{c|cccccc|ccc}
        \toprule
         & \bf target &\bf loss &\bf mask ratio &\bf gap &\bf arch design &\bf sim.\ constrain &\bf
         NN &\bf LP &\bf Ft \\
        \midrule
        rand.\ init. &  &  &  &  &  &  &
        --- & 13.1 & 54.7 \\
        \midrule
        MAE & pixel & MSE & 0.75 & 0 & enc-dec & none &
         30.0 & 70.4 & 88.3 \\
         Naive L-MIM & Latent Feat. & MSE & 0.75 & 0 & enc-dec & none &
         18.8 & 39.3 & 79.7 \\
         & \textcolor{gray}{Latent Feat. } & InfoNCE Patch & \textcolor{lightgray}{0.75} & \textcolor{lightgray}{0} & \textcolor{lightgray}{enc-dec} & \textcolor{lightgray}{none} &
         30.0 & 53.4 & 86.0 \\
          & \textcolor{gray}{Latent Feat. }& \textcolor{gray}{InfoNCE Patch} & 0.9 & \textcolor{lightgray}{0} & \textcolor{lightgray}{enc-dec} & \textcolor{lightgray}{none} &
         53.3 & 72.1 & 89.0 \\
           & \textcolor{gray}{Latent Feat. }& \textcolor{gray}{InfoNCE Patch} & \textcolor{gray}{0.9} & 4 & \textcolor{lightgray}{enc-dec} & \textcolor{lightgray}{none} &
         57.4 & 72.0 & 89.0 \\
         
          & \textcolor{gray}{Latent Feat. }& \textcolor{gray}{InfoNCE Patch} & \textcolor{gray}{0.9} & \textcolor{gray}{4} & enc-proj-xdec &  \textcolor{lightgray}{none} &
         59.9 & 74.0 & 88.7 \\
         
         \method{} & Latent Feat. & InfoNCE Patch & 0.9 & 4 & enc-proj-xdec & 0.1$\times$ (0.25$\rightarrow$0.75) &
        \bf 63.3 & \bf 77.3 & \bf 89.9 \\
         
        \bottomrule
    \end{tabular}
}

%
%
%
%
%
%
%
\subsection{\label{sec:challenge1}Challenge 1: Latent Target Optimization}
Representation collapse in latent MIM is consistent with the findings of negative-free contrastive learning methods like BYOL~\cite{byol}. Unsurprisingly, because both the online and target encoders are trained to minimize the discrepancy between their representations, they can easily lead to a degenerate solution where all images are mapped to the same latent. Common strategies to prevent collapse involve introducing asymmetries between encoders and detaching the target encoder from gradient computation~\cite{byol}. We explore three strategies: 
\begin{description}
    \item[Stand-alone target encoder.] The first strategy is to treat the target encoder independently of the online encoder, without weight sharing.
    \item[Weight-sharing with stop gradient.] Alternatively, we can use a siamese architecture with shared weights, but avoid using the target's gradients to update the online encoder.
    \item[Momentum targets.] Using Momentum encoder~\cite{tarvainen2017mean} is also a common strategy to create asymmetries, and enhance the target encoder simultaneously. As $f(\cdot)$ evolves, the exponential moving average of its weights is tracked and used as the weights of the momentum encoder$\bar{\theta} \leftarrow m\theta+(1-m)\bar{\theta}$. 
\end{description} 

    

\begin{table}[t!]
    \centering
    \begin{minipage}{0.37\linewidth}
        \centering
        \caption{\label{tab:challenge1}{\bf Target encoder optimization.} In addition to downstream performance, we report the average pairwise cosine similarity between (mean pooled) latents $Z$.}
    \end{minipage}
    \hfill
    \begin{minipage}{0.6\linewidth}
        \centering
        \resizebox{\linewidth}{!}{
        \begin{tabular}{ccccc}
            \toprule
            \bf Method & 
            \bf Sim &
            \bf NN & \bf LP & \bf Ft \\
            \midrule
            Naive Latent-MIM & 1.00 & 0.9 & 7.9 & nan \\
            No weight sharing & 0.96 & 3.4 & 7.6 & 44.4 \\
            Weight sharing \& stop-grad & 0.99 & 11.3 & 26.6 & 56.9 \\
            Momentum targets &  \bf 0.50 & 18.8 & 39.3 & 79.7 \\
            \midrule
            MAE~\cite{he2022masked} & 0.67 &  \bf 30.0 &  \bf 70.4 &  \bf 88.3 \\
            \bottomrule
        \end{tabular}}
    \end{minipage}
\end{table}

\paragraph{Study Results} We report the performance of the various models in \cref{tab:challenge1}. To assess representation collapse, we computed the pairwise similarity between samples in the latent space. As expected, the naive latent-MIM model collapses, with all samples mapped to the same latent (similarity of 1.0).
As indicated by the lower cosine similarities, all three strategies help prevent full representation collapse, with momentum targets consistently outperforming the other strategies on downstream classification tasks. 
However, none of the strategies are able to match the performance of MAE, suggesting that latent target optimization is not the only factor hampering the success of latent MIM.

\subsection{\label{sec:challenge2}Challenge 2: Reconstruction Objective for Latent MIM}
In pixel-based MIM, patch reconstruction is enforced by minimizing the mean squared error (MSE) between the decoder's output and the pixel intensities at the target locations. In order to accurately reconstruct the target pixels from a limited visible context, the model is encouraged to learn representations that capture both global semantics as well as patch-specific information. However, since in Latent MIM, the targets are the learned latent representations, we hypothesize that direct reconstruction objectives can also contribute to the optimization challenges, as there are no negative samples to stabilize the learning process. Thus, we investigate the impact of different reconstruction objectives on the effectiveness of Latent MIM.

\paragraph{Direct reconstruction}
We study three loss functions that directly minimize the discrepancy between predicted and target representations, namely MSE, L1, and Huber losses. While the MSE is widely used due to its simplicity and effectiveness, the L1 loss is robust to outliers. The Huber loss combines the best properties of the MSE and L1 losses by being quadratic for small errors and linear for large ones, thus providing a balance between robustness and efficiency. 
Mathematically, the reconstruction losses for the $k$-th target patch are

\begin{eqnarray}
    \Delta_{\textit{L2}}^k = \left\| \hat{\vz}_k - \vz_k \right\|_2^2,\quad
    \Delta_{\textit{L1}}^k = \left\| \hat{\vz}_k - \vz_k \right\|_1,\\
    \Delta_{\textit{Huber}}^k = 
        \begin{cases} 
            \frac{1}{2}\Delta_{\textit{L2}}^k & \text{if}\ \Delta_{\textit{L2}} < \delta^2 \\
            \delta \cdot \left(\Delta_{\textit{L1}}^k - \delta/2\right) & \text{otherwise}.
        \end{cases}
    \label{eq:l2-l1-huber}
\end{eqnarray}

\paragraph{Patch discrimination}
The main drawback of direct reconstruction for Latent MIM is its inability to explicitly incentivize the model to learn diverse representations across the image. This is unlike in pixel-based MIM where the pixel intensities are guaranteed to vary across the image. To circumvent this limitation, we propose a \textit{patch discrimination} objective, where the model is trained to distinguish between target patches using an InfoNCE loss~\cite{cpc}. Specifically, for each target patch $k$, the predicted latent $\hat{\vz}_k$ is contrasted with the latents of all target patches $\vz_l$
\renewcommand{\exp}[1]{\mbox{exp}\left({#1}\right)}
\renewcommand{\sim}[2]{\mathtt{sim}\left({#1},{#2}\right)}

\begin{equation}
    \Delta_\textit{PatchDisc}^k = -\tau \log
    \frac{\exp{-\tfrac{1}{\tau} \sim{\hat{\vz}_k}{\vz_k}}}{
         \sum_{l\in\gT}\exp{-\tfrac{1}{\tau} \sim{\hat{\vz}_k}{\vz_l}}}
    ,\quad
    \mathtt{sim}(\hat{\vz}, \vz)=\frac{\hat{\vz}^T\vz}{\|\hat{\vz}\|\|\vz\|}
    \label{eq:infonce-patch}
\end{equation}

\noindent where $\tau$ is a temperature hyper-parameter.
\footnote{The patch discrimination loss has no batch dependencies, as the negative samples are derived from the image itself.}
To minimize this loss, the model must not only align the predicted and target latents accurately but also ensure sufficient diversity among the latents within the image.


\begin{wraptable}[11]{r}{0.4\linewidth}
    \vspace{-1em}
    \resizebox{\linewidth}{!}{\loss}
    \caption{\label{tab:challenge2} {\bf Loss Impact.}}
\end{wraptable}


\paragraph{Study Results}
We compare each of the aforementioned loss functions for latent reconstruction. Building on the findings of \cref{sec:challenge1}, targets are computed from a momentum encoder. 
The results shown in \cref{tab:challenge2} indicate that, although still insufficient for effective representation learning through latent MIM, the patch discrimination loss (PatchDisc) can learn better representations than with direct reconstruction losses. In particular, we highlight the significant improvements in the nearest neighbor and linear probing accuracy, which are more sensitive to the quality of the learned representations. 

\subsection{\label{sec:challenge3}Challenge 3: Semantic Correlation between Nearby Patches}
Image content displays high correlation within proximate regions. This can render mask reconstruction a trivial task, as the model can interpolate missing information from nearby visible patches. To counter this, pixel-based MIM masks a substantial portion of the image (up to 75\%)~\cite{he2022masked}. 
Latent representations, which are expected to encapsulate high-level semantics, exhibit even higher correlations across patches compared to their corresponding pixels. This correlation potentially undermines the effectiveness of the task for representation learning.

\paragraph{High mask ratio.} Latent MIM also benefits from \textit{high mask ratios}. Beyond reducing patch correlation and enhancing representation learning, this strategy also enables faster training and a lower GPU memory footprint. However, it has to keep enough visible patches to capture critical features within the image.

\begin{figure}[t]
    \centering
    \begin{minipage}{0.42\linewidth}
        \centering
        \includegraphics[width=\textwidth]{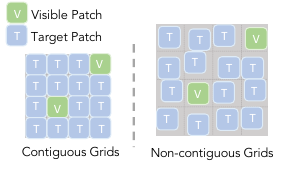} 
        \caption{\label{fig:patchgeneration} Patch Generation Strategies. Left: Masking contiguous grids. Right: Non-contiguous stochastic masking.}
    \end{minipage}
    \hfill
    \begin{minipage}{0.52\linewidth}
        \centering
        \resizebox{0.85\linewidth}{!}{\combinedcorrelation}
        \captionof{table}{\label{tab:combined} Improvement from reducing semantic correlation mitigation with strategies of high mask ratio, non-contiguous stochastic maskin and patch-wise similarity constraint.}
    \end{minipage}
\end{figure}

\paragraph{Non-contiguous grids.} To further minimize the correlation between visible and target patches without reducing the amount of visible cues, we experiment with stochastic non-contiguous grids (\cref{fig:patchgeneration}). This strategy increases the distance between patches by separating each patch from its neighbors by a random number of unused pixels. 
Specifically, let $P$ represent the patch size and $G$ the average gap between consecutive patches. Stochastic non-continuous grids can be conveniently generated by initially splitting the image into a regular grid of patches, each of size $(P+G)\times(P+G)$, and then extracting a $P\times P$ patch at random from each grid location.

\paragraph{Patch Similarity Constraints} While the previous strategies reduce correlation by refining the patch selection process, we can also impose explicit constraints to avoid correlation in the latent space. This is especially important since, as highlighted in the previous challenge (\cref{tab:challenge2}), patch representations tend to cluster together when trained with no constraints. To counteract this, we impose a constraint on both the visible and predicted latents, $Z_\gV$ and $\hat{Z}_\gT$,
\begin{equation}
    \gR = {\left(\gamma - \E_{i,j\in\gT} \left[ \sim{\hat{\vz}_i}{\hat{\vz}_j} \right] \right)}^2 
    + {\left(\gamma - \E_{i,j\in\gV} \left[ \sim{\vz_i}{\vz_j} \right] \right)}^2,
    \label{eq:patch-sim}
\end{equation}
where $\gamma$ is a predefined desired inter-patch similarity.

%
%
%
%
%

\paragraph{Study Results}
Once again, we build on the findings of the previous challenge and use patch discrimination for latent reconstruction. 
\cref{tab:combined} shows the incremental improvement in representation quality from each different strategy for mitigating semantic correlation.  Latent MIM shows substantial improvements from even higher mask ratios than pixel-based MIM, with the optimal ratio being 90\%. Exceeding 90\% masking is counterproductive, as not enough visible patches would be available to capture critical features within the image. 
Using non-contiguous stochastic masking with gap=$4$ can further improve the representation quality by lowering the spatial redundancy while keeping enough visible information. Finally, adding a regularizer that explicitly constrains the mean similarity among patches also enhances performance. We put the ablation of each strategy and more exploration in \cref{sec:supp_abalation}.

\subsection{\label{sec:challenge4}Challenge 4: Decoder Design for Latent Reconstruction}

The final important component of Latent MIM is the decoder, responsible for predicting target representations from visible patches. The decoder design is crucial for the model's ability to effectively utilize the high-level semantics extracted from the encoder. We explore the impact of different decoder architectures.

\begin{figure}[t!]
    \centering
    \includegraphics[width=\linewidth]{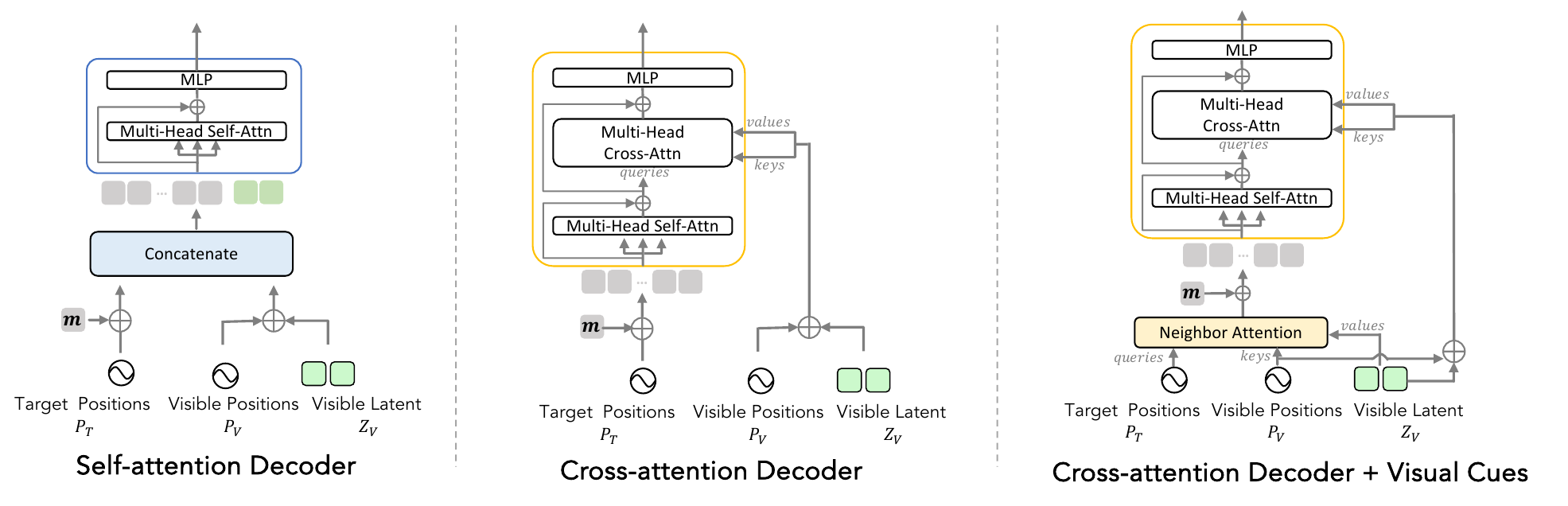}
    \caption{\label{fig:decoder}\textbf{Comparison of Three Different Decoder Designs.} 
    Self-attention decoder is commonly used for low-level MIM models. Cross-attention decoder provides direct conditioning at each layer on the visible latents.}
\end{figure}

\paragraph{Self-attention decoder} Pixel-based MIM models employ a self-attention transformer to predict the target pixels $X_\gT$. To accomplish this, the decoder receives two sets of inputs: the latents of visible patches $Z_\gV$ and a sequence of learnable mask tokens $\vm$, marked by the fixed SinCos positional embeddings $\vp$ of their corresponding locations (\ie, $\vm+\vp_t\ \forall t\in\gT$ and  $\vz_v+\vp_v\  \forall v\in\gV$). After processing this sequence through a series of self-attention blocks, the decoder outputs the target representations using a linear head.

\paragraph{Cross-attention decoder} Unlike pixel-based MIM, in Latent MIM the targets $Z_\gT$ are representations with a similar level of abstraction than the features obtained from the encoder $Z_\gV$. Thus, the decoder should be able to condition on these visible representations $Z_\gV$ more directly. While self-attention only conditions once $Z_\gT$ through the input sequence, cross-attention allows the decoder to condition on $Z_\gV$ at every layer. A standard cross-attention architecture~\cite{vaswani2017attention} with alternating self-attention, cross-attention, and feed-forward MLP blocks is used to update the prediction tokens $\vm+\vp_t\ \forall t\in\gT$.

\paragraph{Visual cues from neighboring visible patches} 
As discussed in (\cref{sec:challenge3}), neighboring patch latents can be highly correlated. To prevent the decoder from focusing excessively on interpolating between these patches, we embed visual cues directly into its input sequence. This allows the decoder to better focus on more fine-grained spatial reasoning.
Specifically, with $P_\gV$ and $P_\gT$ as the positional embeddings for the visible and target patches, respectively, we initialize the prediction tokens as $\vm_i = \vm +\vp_t + \textit{Softmax}_t\left(P_\gT {P_\gV)}^T\right) Z_\gV$ at each target location $i\in\gT$. This setup equips the mask tokens $M_\gT=\lbrace\vm_i\rbrace_{i\in\gT}$ with a weighted blend of latents from the nearest visible patches, providing precise location and visual cues right from the start.

\paragraph{Latent Projector} Following contrastive learning methods, we experiment with non-linear projection heads $h(\cdot)$ to prepare the latents from visible patches $Z_\gV$ for the decoder. Through a series of experiments with various non-linear projectors $h$, we found that a simple multi-layer perceptron (MLP) with three layers, GELU activations, and layer normalization yielded the most effective representations.

%
%
%
%

    

\begin{table}[t!]
    \centering
    \begin{minipage}{0.32\linewidth}
        \centering
        \caption{\label{tab:combined-decoder}{\bf Comparison between different decoder designs}. Other configurations follow the optimal settings found in previous sections.}
    \end{minipage}
    \begin{minipage}{0.65\linewidth}
        \centering
        \resizebox{\linewidth}{!}{\combineddecoder}
    \end{minipage}
\end{table}

\paragraph{Study Results}
\cref{tab:combined-decoder} assesses the impact of different decoder designs on the quality of the learned representations. 
Pixel-based MIM models, such as MAE, employ a lightweight self-attention decoder with a depth of 8 layers. Our findings underscore the necessity for a distinct decoder design for Latent MIM, given the different nature of the decoder's task.
Specifically, we found that cross-attention provides a better mechanism for conditioning the reconstruction on the visible patches, leading to superior performance in all downstream tasks, especially when initializing the prediction tokens with visual cues from neighboring visible patches and when combined with a non-linear projection head to process the input visible patches. 
Furthermore, since both the inputs and targets are high-level, the Latent MIM decoder can be remarkably lightweight. Optimal performance is achieved using a shallow (3-layer) transformer, considerably smaller than the 8-layer MAE decoder.
When combined, these strategies allow the decoder to better utilize the high-level semantic features extracted from the encoder, surpassing the performance of pixel-based MIM models in all downstream tasks.

\section{Scaling to ImageNet-1k\label{sec:in1k}}
The previous section provided a detailed analysis of the optimization challenges and design decisions for Latent MIM on a medium-sized dataset. In this section, we show the scalability of Latent MIM to larger datasets, specifically ImageNet-1k, and compare it to prior work. We also highlight the strongly localized semantics learned by Latent MIM by evaluating the trained model on unsupervised scene segmentation, video object segmentation, and few-shot transfer learning.

\paragraph{Implementation}
We scaled up pretraining to ImageNet-1k using the optimal Latent MIM configuration from \cref{sec:challenges}. The model is pretrained for 800 epochs using the Adam optimizer. Full implementation details are in the \cref{sec:implementation}.

\newcommand{\infull}{
\begin{tabular}{lccccc}
    \toprule
    \bf Method & \bf Epochs & 
    \bf NN & \bf Lin & \bf Ft \\
    \midrule
    \mc{3}{l}{\textit{\small $\triangleright$ Low-level MIM}}  & & \\
    {MAE~\cite{he2022masked}} & {1600} & {12.2} & {67.8} & {83.6} \\
    {MaskFeat~\cite{wei2022masked}}  & {300} & {-} & {-} & {83.6} \\
    {SimMIM~\cite{xie2022simmim}}  & {800} & {-} & {56.7} & {83.8} \\
    \midrule
    \multicolumn{3}{l}{\textit{\small $\triangleright$ MIM using frozen pretrained target}} & &  \\
    {BeiT~\cite{beit}} & {300} & {-} & {56.7} & {82.9} \\
    {PeCo~\cite{dong2023peco}} & {800} & {-} & {-} & {\bf 84.5} \\
    \midrule
    \multicolumn{3}{l}{\textit{\small $\triangleright$ Latent MIM}} & & \\
    ConMIM~\cite{yi2022masked} & 800 & - & 39.3 & 83.7\\
    data2vec~\cite{baevski2022data2vec} & 800 & 25.7 & 60.3 & 84.2 \\
    \rowcolor{blue!10} Latent-MIM (Ours) & 800 & \bf 50.1 & \bf 72.0 & 83.0 \\
    \midrule
    \midrule
    \multicolumn{3}{l}{\textit{\small $\triangleright$ Global Joint Embeddings}} & &  \\
    \color{gray}{MoCo v3~\cite{mocov3}} & \color{gray}{300}$\times$2 & \color{gray}{61.3} & \color{gray}{76.7} & \color{gray}{83.2} \\
    \color{gray}{DINO~\cite{dino}} & \color{gray}{300} & \color{gray}{-} & \color{gray}{78.2} & \color{gray}{-}\\
    \midrule
    \multicolumn{3}{l}{\textit{\small $\triangleright$ Hybrid MIM}} & & \\
    \color{gray}{iBoT~\cite{zhou2021ibot}}  & \color{gray}{1600} & \color{gray}{-} & \color{gray}{79.5} & \color{gray}{84.0} \\ 
    \color{gray}{SIM} & \color{gray}{1600} & \color{gray}{-} & \color{gray}{76.4} & \color{gray}{83.6} \\ 
    \color{gray}{CAE~\cite{chen2023context}}  & \color{gray}{800} & \color{gray}{-} & \color{gray}{68.6} & \color{gray}{83.8} \\ 
    \bottomrule
\end{tabular}
}

\newcommand{\incls}{
    \begin{tabular}{=l +c +c +c +c }
        \toprule
        \bf Method & \bf Epochs & 
        \bf NN & \bf LP \\ 
        \midrule
        
        \multicolumn{3}{l}{\textit{\small Low-level}}  & & \\
        MAE~\cite{he2022masked} & 1600 & 12.2 & 67.8 \\ 
        SimMIM~\cite{xie2022simmim}  & 800 & - & 56.7  \\ 
        \midrule
        
        
        \multicolumn{3}{l}{\textit{\small Latent}} & & \\
        ConMIM~\cite{yi2022masked} & 800 & - & 39.3  \\ 
        data2vec~\cite{baevski2022data2vec} & 800 & 25.7 & 60.3  \\ 
        \rowcolor{blue!10}
        Latent MIM  & 800 & \bf 50.1 & \bf 72.0  \\ 
        \bottomrule

        
    \end{tabular}
}

\newcommand{\davis}{
    \begin{tabular}{lc|ccc}
        \\\\
        \hline
        \bf Method & \bf  epochs &  \( \mathcal{J}\&\mathcal{F} _m \) & \( \mathcal{J}_m \) & \( \mathcal{F}_m \) \\
        \hline
        MAE  & 1600 & 57.5 & 54.8 & 60.2 \\
        data2vec  & 800 & 28.5 & 27.8 & 29.2 \\
        Latent MIM  & 800 & \bf 65.5 & \bf63.1 & \bf 68.0 \\
        \hline
    \end{tabular}
}

\newcommand{\fewshot}{
        \begin{tabular}{lccccccc}
        \toprule
        \multicolumn{1}{c}{}
        \bf Method &
        \bf \thead{Caltech101\\\cite{fei2004learning}} &  
        \bf \thead{DTD\\\cite{cimpoi14describing}} & 
        \bf \thead{Oxford\\Flowers~\cite{nilsback2008automated}} & 
        \bf \thead{Oxford\\Pets~\cite{parkhi2012cats}} & 
        \bf \thead{Stanford\\Cars~\cite{krause20133d}} &
        \bf \thead{SUN397\\\cite{xiao2016sun}} & 
        \bf \thead{UCF101\\\cite{soomro2012ucf101}} \\
        \midrule
        MAE     & 80.5\std{0.4} & 51.8\std{2.2} & 62.7\std{1.8} & 60.2\std{3.6} & 7.8\std{1.0} & 21.9\std{0.4} & 53.2\std{0.4}\\
        data2vec    & 76.6\std{2.1} & 52.0\std{2.1} & 74.1\std{2.2} & 58.5\std{2.0} & 8.8\std{0.4} & 29.0\std{0.5} & 52.6\std{1.2}\\
        Latent MIM & {\bf 89.2}\std{1.0} &  {\bf 55.9}\std{2.0} & {\bf 84}\std{1.0} & {\bf79.8}\std{2.6} & {\bf 16.3}\std{2.0} & {\bf48.4}\std{0.6} & {\bf75.8}\std{2.8} \\
        \bottomrule
    \end{tabular}
}

\subsection{ImageNet-1k Classification}
Following~\cite{he2022masked}, we evaluate the learned representations on ImageNet-1K using Nearest Neighbor (NN) and linear probing (LP) protocols. \cref{tab:in1k-cls} compares our Latent MIM with low-level MIM methods and prior latent MIM methods with ViT-B/16. For MAE and data2vec, we use the features (either cls-token or average-pooling) that yield the best performance. 

\begin{wraptable}[11]{r}{0.5\linewidth}
    \vspace{-3.5em}
    \caption{\label{tab:in1k-cls}Top-1 NN and LP classification accuracy on ImageNet.}
    \centering
    \resizebox{\linewidth}{!}{\incls}
\end{wraptable}
Latent MIM surpasses low-level MIM methods by large margins on NN classification (+37.2\% on MAE), and LP (+3.3\% on MAE and +14.4\% on SimMIM), demonstrating the effectiveness of Latent MIM for learning improved semantics from latent masked reconstruction. Our method also outperforms related prior work, data2vec and ConMIM, by large margins.

\subsection{Properties beyond Classification}
The Latent MIM framework is designed to learn localizable and semantically rich representations. We showcase these properties on unsupervised segmentation, semi-supervised video object segmentation, and transfer learning.

\begin{figure}[t]
    \centering    
    \begin{subfigure}[b]{0.3\linewidth}
        \centering
        \includegraphics[width=\linewidth]{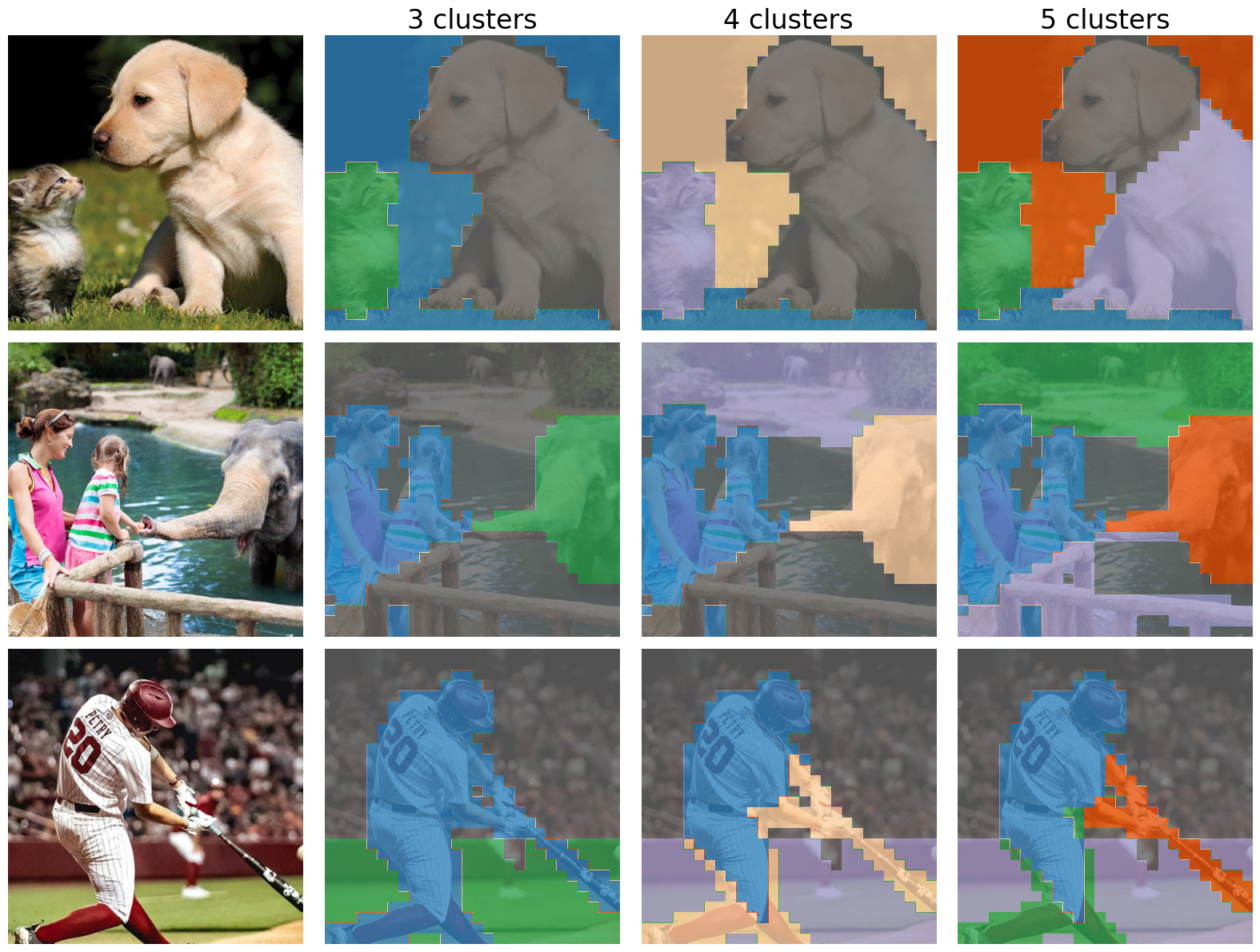}
        \caption{Latent MIM}
    \end{subfigure}\hfill
    \begin{subfigure}[b]{0.3\linewidth}
        \centering
        \includegraphics[width=\linewidth]{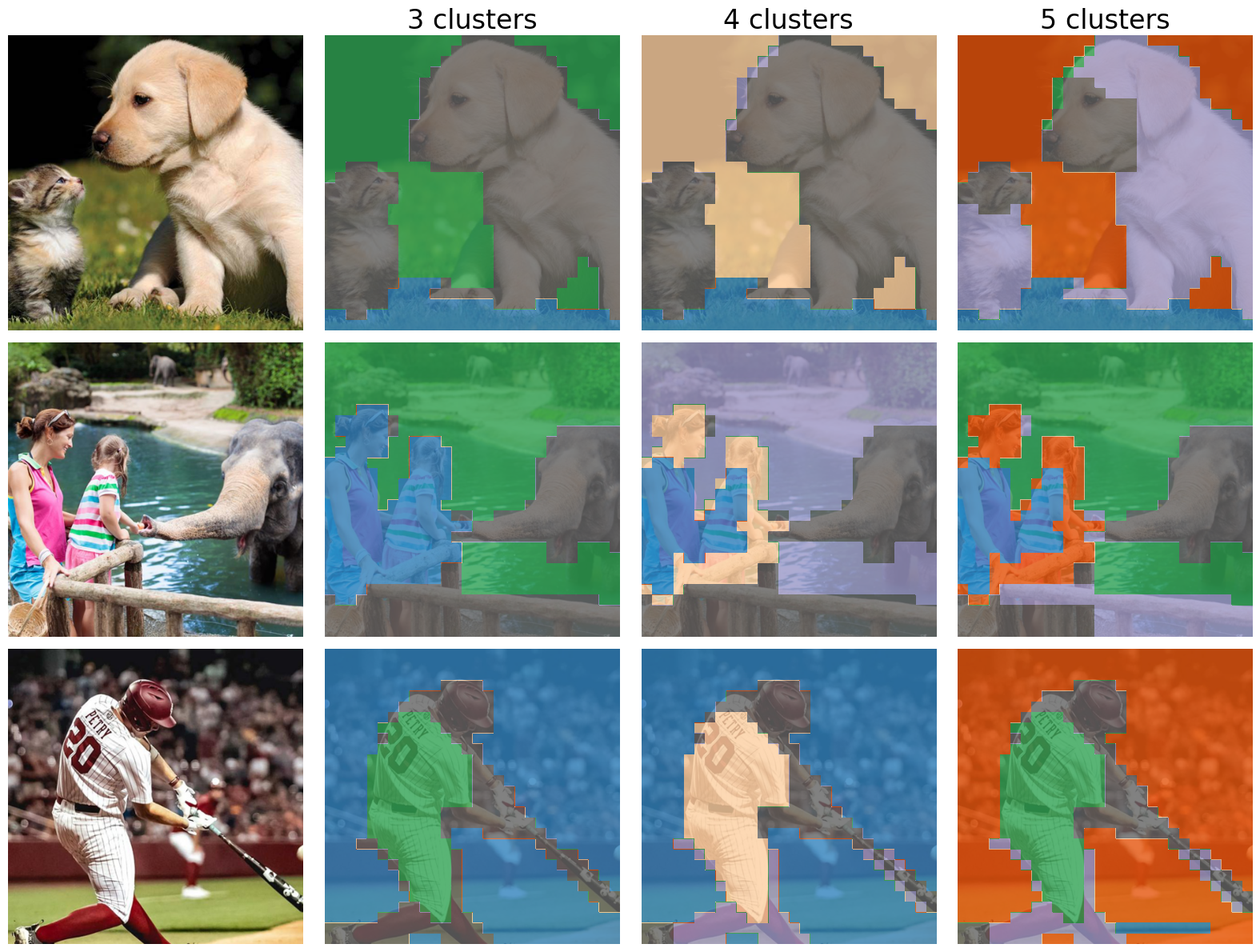}
        \caption{data2vec}
    \end{subfigure} \hfill
    \begin{subfigure}[b]{0.3\linewidth}
        \centering
        \includegraphics[width=\linewidth]{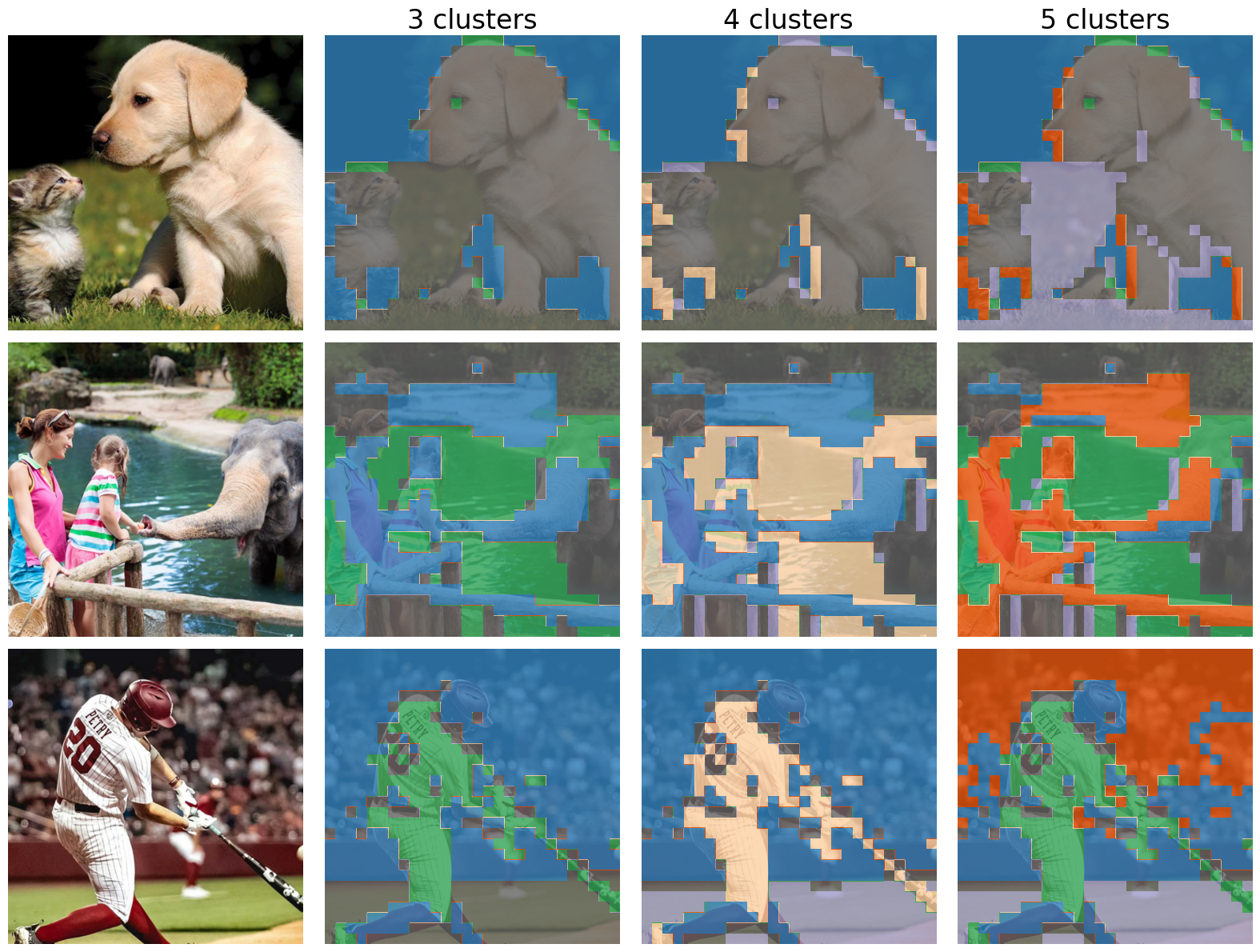}
        \caption{MAE}
    \end{subfigure}
    \caption{\label{fig:segm}\textbf{Unsupervised segmentation maps} by hierarchical clustering of patch representations within each image. Better viewed digitally with zoom.}
\end{figure}

\paragraph{Unsupervised segmentation}
An emerging property of Latent MIM is its capacity for semantic clustering of local representations, which enables impressive segmentation and scene parsing outcomes without the need for supervised fine-tuning. \cref{fig:segm} illustrates the unsupervised segmentation maps generated by hierarchical clustering of patch-level representations. Compared to both lower-level MIM approaches, such as MAE, and earlier latent MIM methods like data2vec, our Latent MIM model learns better semantic and localizable representations. 


\begin{figure}[t]
    \begin{minipage}{0.5\textwidth}
        \centering
        \resizebox{\linewidth}{!}{\davis}
        \captionof{table}{\label{tab:davis}\textbf{Video object segmentation} on DAVIS-2017. $\mathcal{J}$ and  $\mathcal{F}$ quantify region similarity and boundary alignment, respectively.}
    \end{minipage}
    \hfill
    \begin{minipage}{0.45\textwidth}
        \centering
        \includegraphics[width=\linewidth]{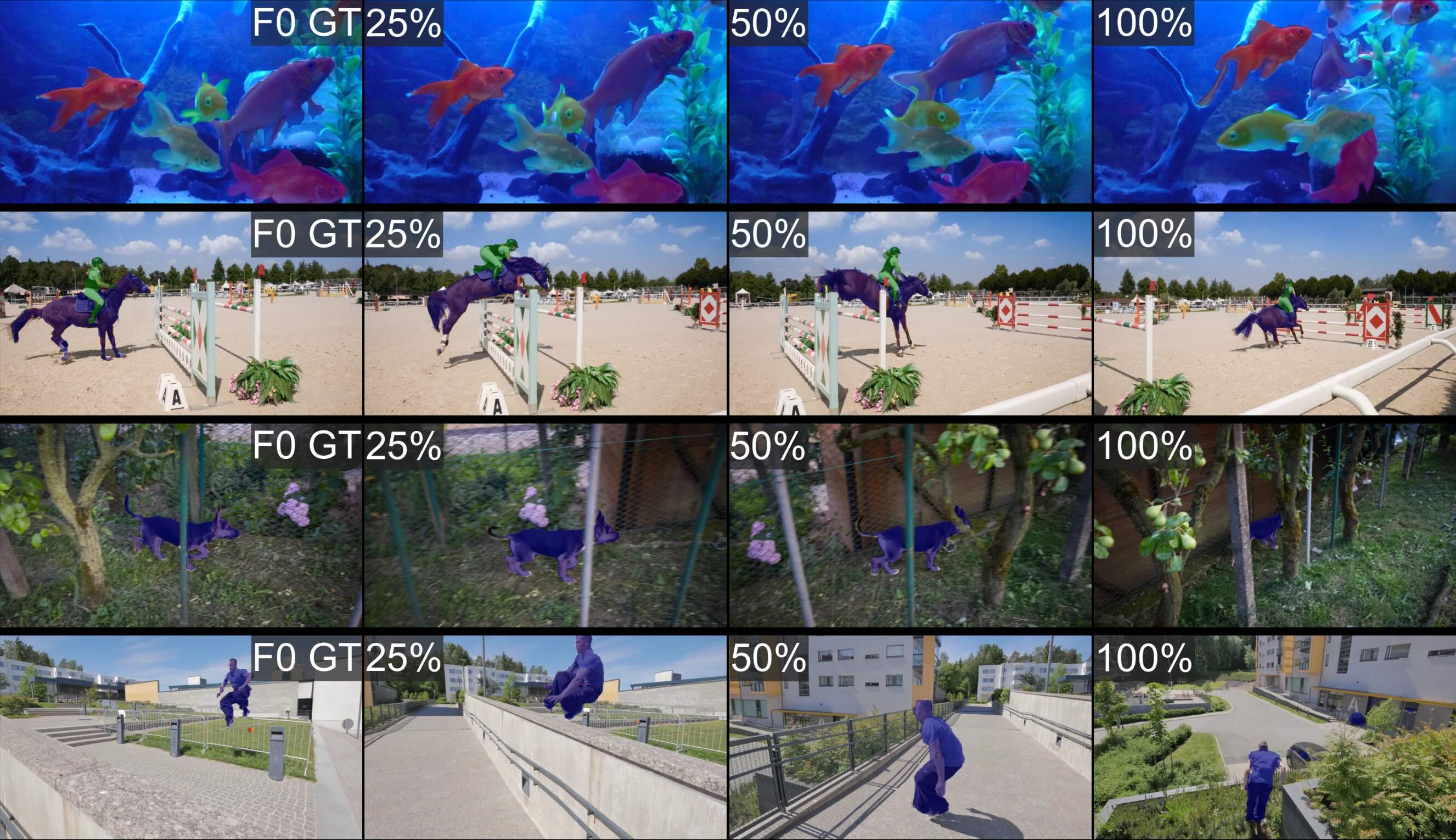}
        \caption{\label{fig:davis}\textbf{Visualization}. Col 1: ground truth; Col 2-3: predictions at 25\%, 50\%, and 100\% timesteps.}
    \end{minipage}
\end{figure}



\paragraph{Video Object Segmentation}
Our Latent MIM model can also maintain both semantic integrity and localization accuracy, even in complex, dynamic video sequences. DAVIS-2017 semi-supervised video object segmentation benchmark evaluates the ability to generate precise object segmentation masks in videos, starting from ground truth masks of the initial frame. We follow the experimental protocol in~\cite{jabri2020space}, which segments scenes through a nearest-neighbor strategy applied between consecutive frames. Crucially, this benchmark tests the robustness and adaptability of the pre-trained features without the need for additional training.~\cref{tab:davis} shows that Latent MIM surpassed both pixel-level and previous Latent MIM models in performance.~\cref{fig:davis} shows sample qualitative results.

\begin{table}[t!]
    \caption{\textbf{Few-shot learning on various datasets.} We report the mean and standard deviation of the top-1 accuracy across 3 different runs. Each run is trained on a set of 16 training images per class. \label{tab:fewshot}}
    \centering
    \resizebox{\linewidth}{!}{\fewshot}
\end{table}
\paragraph{Few-shot transfer learning} 
To assess generalization beyond ImageNet-1k, we transfer the learned representations to a variety of datasets. Following the setting of~\cite{zhou2022conditional}, we perform 16-shot transfer learning experiments by training a linear classifier on top of frozen representations. 
All models are trained using SGD with a momentum of 0.9, a learning rate of 0.1, a batch size of 128, and updated for a total of 2500 iterations. 
\cref{tab:fewshot} demonstrates the superior performance of Latent MIM over MAE and data2vec across all datasets, highlighting its robustness, versatility, and potential for various recognition tasks.

\section{Conclusion}\label{sec:conclusion}
We identified and addressed the key training challenges in Latent MIM, demonstrating its capacity to generate spatially diverse, high-level semantic representations.  This is evidenced by significant improvements in nearest neighbor and linear probe evaluation on ImageNet, fewshot transfer learning, as well as in segmentation tasks requiring minimal or no supervision. We hope this work will inspire further exploration into Latent MIM for learning fine-grained semantics without human supervision.

\bibliographystyle{style-eccv/splncs04}
\bibliography{refs}

\newcommand{\newgap}{
\begin{tabular}{cccc}
    \toprule
    \bf g & 
    \bf NN & \bf Lin & \bf Ft \\
    \midrule
    0 & 27.6 & 51.8 & 83.3 \\
    2 & 44.2 & 66.5 & 87.1 \\
    \rowcolor{gray!10} 4 & 63.3 & \bf 77.3 & \bf 89.9  \\
    6 & 62.2 & 76.5 & 89.0 \\
    8 & \bf 63.6 & 75.9 & 89.2 \\
    \bottomrule
\end{tabular}
}

\newcommand{\newmaskratio}{
    \begin{tabular}{cccc}
        \toprule
        \bf mask ratio & 
        \bf NN & \bf Lin & \bf Ft \\
        \midrule
        0.5 & 7.6 & 26.3 &  71.4\\
        0.75 & 44.5 & 71.3 & 89.7 \\
        0.85 & 61.7 & 76.9 &  89.8\\
        \rowcolor{gray!10} 0.9 & \bf 63.3 & \bf 77.3 & \bf 89.9  \\
        0.95 & 51.3 & 71.0 &  87.8 \\
        \bottomrule
    \end{tabular}
}

\newcommand{\newprojector}{
    \begin{tabular}{cccc}
        \toprule
        \bf projector & 
        \bf NN & \bf Lin & \bf Ft \\
        \midrule
         none & 57.8 & 73.0 & 88.7 \\ 
         \rowcolor{gray!10} mlp & \bf 63.3 & \bf 77.3 & \bf 89.9  \\
        \bottomrule
    \end{tabular}
}

\newcommand{\newmodeldesign}{
    \begin{tabular}{cccc}
        \toprule
        \bf decoder & 
        \bf NN & \bf Lin & \bf Ft \\
        \midrule
        Self-attn & 59.7 & 76.8 & 89.4 \\
        \rowcolor{gray!10} Cross w/ Visual Cues & \bf 63.3 & \bf 77.3 & \bf 89.9  \\
        \bottomrule
    \end{tabular}
    }

\newcommand{\newloss}{
    \begin{tabular}{cccc}
        \toprule
        \bf loss & 
        \bf NN & \bf Lin & \bf Ft \\
        \midrule
        MSE & 58.4 & 71.6 & 87.4 \\
        \rowcolor{gray!10} PatchDisc & \bf 63.3 & \bf 77.3 & \bf 89.9 \\
        \bottomrule
    \end{tabular}
}

\newcommand{\newavgsimconstrain}{
    \begin{tabular}{cccc}
        \toprule
        \bf Sim & 
        \bf NN & \bf Lin & \bf Ft \\
        \midrule
        \xmark & 59.9 & 74.0 & 88.7  \\ 
        \rowcolor{gray!10} \cmark & \bf 63.3 & \bf 77.3 & \bf 89.9 \\ 
        \bottomrule
    \end{tabular}
}

\newcommand{\newdepth}{
    \begin{tabular}{cc@{\hspace{2em}}ccc}
        \toprule
        \bf \thead{Target\\depth } & \bf \thead{Decoder\\depth } 
        & \bf NN & \bf LP & \bf Ft \\
        \midrule
        0 & 7 & 44.0 & 67.4 & 86.4 \\
        3 & 5 & 52.7 & 74.3 & 88.7 \\
        6 & 4 & 52.7 & 74.8 & 88.9 \\
        9 & 4 & 53.6 & 75.8 & 89.4 \\
        12& 3 & \bf 63.3 & \bf 77.3 & \bf 89.9  \\
        \bottomrule
    \end{tabular}
}

\newcommand{\maintraincfg}{
    \begin{tabular}{rcc}
        \toprule
        \bf Config & \bf Value (Main, Sec 3) &  \bf Value (IN1K, Sec 4) \\
        \midrule
        Backbone & ViT-B/16 \\
        Dataset & ImageNet-100 & ImageNet-1k \\
        Optimizer & AdamW \\
        Epochs & 300 & 800 \\
        Batch size & 1024 & 4096 \\
        LR Schedule & Cosine Decay w/ Warmup \\
        Base LR & 1.5e-4 \\
        Warmup Epochs & 30 & 40 \\
        Weight Decay & 0.05 \\
        Optimizer Momentum & $\beta_1=0.9$, $\beta_2=0.95$ \\
        Target Encoder Momentum & $0.99\rightarrow 1.0$\\
        Data Augmentations & \begin{tabular}{@{}l}RandomHorizontalFlip\\RandomResizedCrop\end{tabular} \\
        \bottomrule
    \end{tabular}
}

\newcommand{\mainlpcfg}{
    \begin{tabular}{rcc}
        \toprule
        \bf Config & \bf Value (Main, Sec 3) &  \bf Value (IN1K, Sec 4) \\
        \midrule
        Dataset & ImageNet-100 & ImageNet-1k \\
        Optimizer & LARS \\
        Epochs & 20 & 90 \\
        Batch size & 1024 & 16384 \\
        LR Schedule & Cosine Decay w/ Warmup \\
        Base LR & 0.5 & 0.1 \\
        Warmup Epochs & 2 & 10 \\
        Weight Decay & 0. \\
        Optimizer Momentum & 0.9 \\
        Data Augmentations & \begin{tabular}{@{}l}RandomHorizontalFlip\\RandomResizedCrop\end{tabular} \\
        \bottomrule
    \end{tabular}
}

\newcommand{\mainftcfg}{
    \begin{tabular}{rcc}
        \toprule
        \bf Config & \bf Value (Main, Sec 3) &  \bf Value (IN1K, Sec 4) \\
        \midrule
        Dataset & ImageNet-100 & ImageNet-1k \\
        Optimizer & AdamW \\
        Epochs & 50 & 100 \\
        Batch size & 1024 \\
        LR Schedule & Cosine Decay w/ Warmup \\
        Base LR & 1e-3 & 5e-4 \\
        Warmup Epochs & 5 & 10 \\
        Weight Decay & 0.05 \\
        Optimizer Momentum & $\beta_1=0.9$, $\beta_2=0.999$ \\
        Data Augmentations & RandAug(9,0.5) \\
        Label Smoothing & 0.1 \\
        Mix-Up & 0.8 \\
        Cut-Mix & 1.0 \\
        Drop Path & 0.1 \\
        \bottomrule
    \end{tabular}
}

\appendix
\section*{Appendix}
\section{Implementation Details \label{sec:implementation}}

\subsection{Optimization details}
To ensure reproducibility, we provide detailed implementation details for all our experiments in \cref{tab:configs}. The table shows the hyperparameters used for pre-training and the linear probing and finetuning downstream tasks, both for the main study (Section 3 of main text) and on the ImageNet-1k dataset (Section 4 of main text). The configurations are consistent with the ones used in prior work~\cite{he2022masked}.

\begin{table}[p!]
    \centering
    \caption{\label{tab:configs}Implementation details used for the main study (Section 3) and on the ImageNet-1k dataset (Section 4) if different.}
    
    \begin{subtable}{0.8\textwidth}
        \centering
        \caption{Pretraining config.\label{tab:main-pretrain-cfg}}
        
        \scalebox{0.75}{\maintraincfg}
    \end{subtable}\\
    \begin{subtable}{0.8\textwidth}
        \centering
        \caption{Linear probing config.\label{tab:main-lp-cfg}}
        
        \scalebox{0.75}{\mainlpcfg}
    \end{subtable}\\
    \begin{subtable}{0.8\textwidth}
        \centering
        \caption{Finetuning config.\label{tab:main-ft-cfg}}
        
        \scalebox{0.75}{\mainftcfg}
    \end{subtable}
\end{table}

\subsection{Latent MIM configuration used for ImageNet-1k (Section 4)}
We scaled up pre-training to ImageNet-1k using the optimal Latent MIM configuration discussed in Section 3 of the main text. 
We use the standard ViT-B/16~\cite{vits} as the online and momentum encoder and a 3-layer cross-attention decoder with visual cues described in Section 3 of the main text. The projector is a 3-layer MLP with 4096 hidden dimensions, GELU activation, and layer-norm. We use a 90\% mask ratio to mask the 14$\times$14 grid in a non-contiguous way with gap $G=4$, resulting in 20 visible and 176 target patches per sample. 
The model is trained with a Patch Discrimination loss together with similarity constraints on both $Z_\gV$ and $\hat{Z}_\gT$ with a 0.1 coefficient where the target similarity is updated from 0.75 to 0.25 following the cosine schedule.
For both linear probing and finetuning, we also experimented with the pooling method used for feature extraction before classification. In addition to the standard max pooling, average pooling and the use of the CLS token, we apply a pooling method in between max and average pooling, which averages the top $k$ feature values. Empirical results indicate that setting $k=10$ yields optimal performance.
\section{Supplementary Analysis\label{sec:supp_abalation}}

\subsection{Comparison with I-JPEA}
\begin{figure}[t]
    \centering
    \begin{minipage}[b]{0.36\textwidth}
        \centering
        \includegraphics[width=\textwidth]{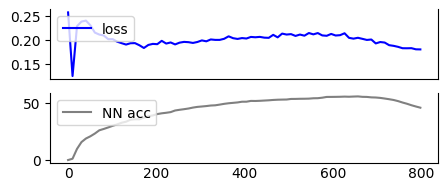} 
        \caption{I-JPEA training.}
        \label{fig:ijepa_vitb14_curve}
    \end{minipage}
    \begin{minipage}[b]{0.6\textwidth}
        \centering
         \resizebox{\linewidth}{!}{
        \begin{tabular}{lc|ccc}
        \hline
        \bf Method & \bf Backbone & \( \mathcal{J}\&\mathcal{F} _m \) & \( \mathcal{J}_m \) & \( \mathcal{F}_m \) \\
        \hline
        I-JEPA & ViT-H/14 &62.5 & 60.6 & 64.4 \\
        Latent MIM & ViT-B/16 & \bf 65.5 & \bf63.1 & \bf 68.0 \\
        \hline
        \end{tabular}}
        \captionof{table}{Comparison to I-JEPA on Davis-2017.\label{tab:ijepa-davis}}
    \end{minipage}
\end{figure}

\begin{figure}[t]
    \centering    
     \begin{subfigure}[b]{0.65\linewidth}
        \includegraphics[width=0.9\linewidth]{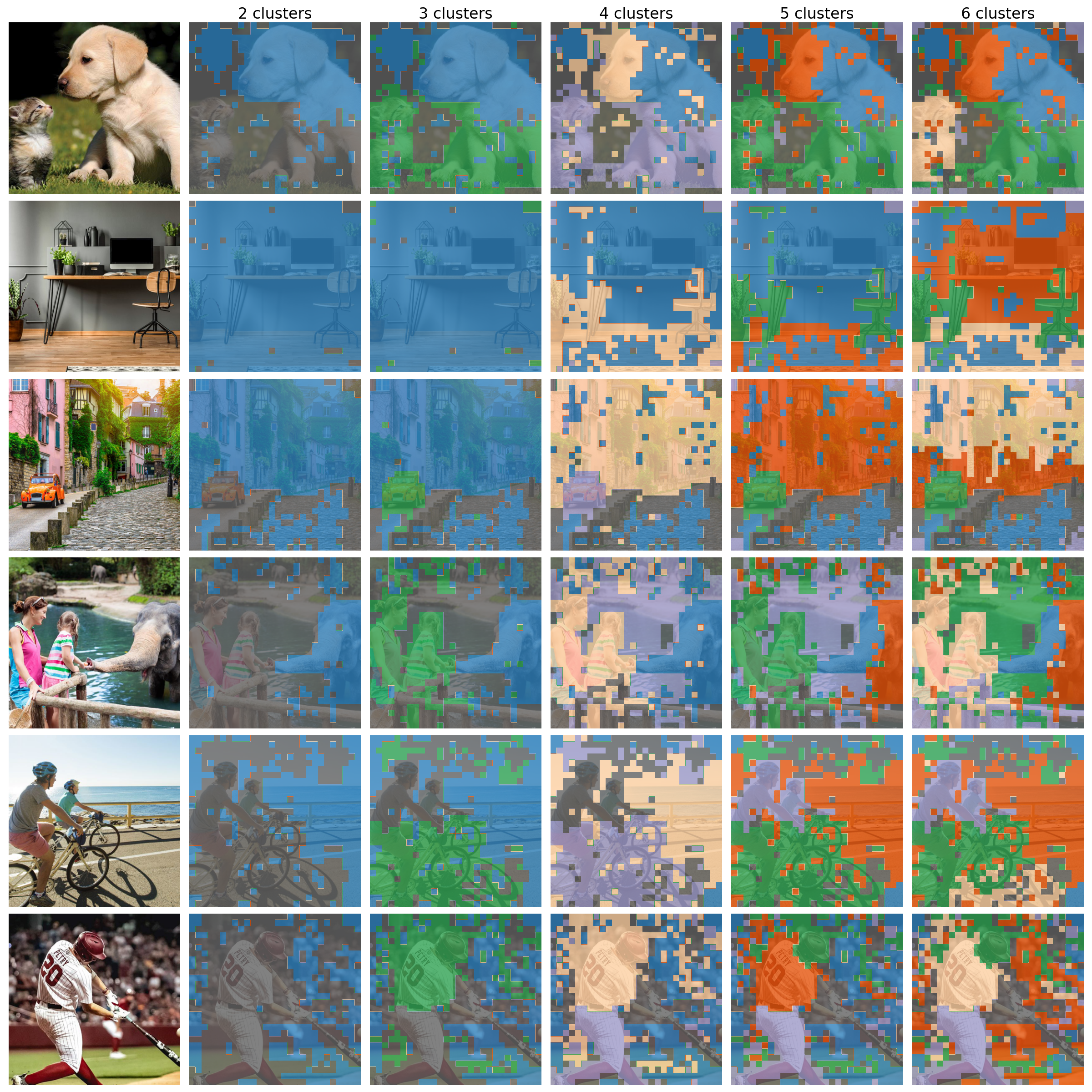}
        \caption{I-JEPA (ViT-H/14)}
    \end{subfigure}\hfill
     \begin{subfigure}[b]{0.35\linewidth}
        \centering  
        \includegraphics[width=0.28\linewidth]{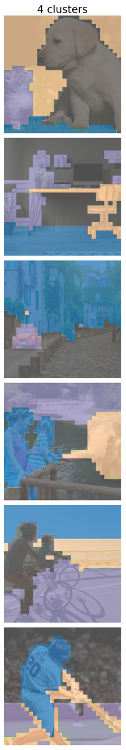}
        \caption{Ours(ViT-B/16)}
    \end{subfigure}
    \caption{\label{fig:ijepa-seg} Unsupervised segmentation maps}
\end{figure}

I-JEPA is a prior exploration in pure latent MIM and has shown promising results, but it still shows signs of training instability. \cref{fig:ijepa_vitb14_curve} shows I-JEPA training curve trained for 800 epochs with ViT-B on ImageNet-1K. Note the clear drop in performance at the end of the training, which likely occurs due to the direct MSE patch reconstruction loss, which we show to be vulnerable to degenerate solutions. We also found that our method obtains better-localized semantics than I-JEPA. We compared our model against the pre-trained I-JEPA VIT-H/14 on video object propagation (\cref{tab:ijepa-davis}) and unsupervised segmentation (\cref{fig:ijepa-seg}). Our model outperformed I-JEPA even with a smaller backbone. These enhancements are due to our emphasis on localized semantics, which is a key benefit of Latent MIM. Our study identified methods to stabilize latent MIM training without sacrificing localized semantics. 

\subsection{Latent MIM: From low-level to high-level latents.}

To assess the effectiveness of the learned representations with different types of targets, we conducted experiments using momentum encoders of varying depths in our training process, ranging from a depth of 0 (i.e., pixel reconstruction) to the full momentum encoder of depth 12. We also observed that the choice of target depth significantly affected the depth of the optimal decoder. Therefore, for each target depth, we explored different decoder depths and reported the optimal configuration. These experiments were conducted using the experimental setup of Section 3 in the main paper, with the optimal configuration found after Challenge 4. \cref{tab:depth} compares the downstream performance of the various models. First, as expected, pixel-level and low-level targets require larger decoders, since the gap between the encoder representations and target representations is larger. Second, as the target momentum encoder becomes shallower, we observe that the downstream performance of the learned representations decreases sharply across all evaluation metrics. These results show that the reconstruction of local high-level targets can be highly beneficial, motivating the Latent MIM framework.

\begin{table}[h]
    \centering
    \caption{\label{tab:depth}Downstream performance of Latent MIM models on IN100 dataset using momentum encoders and decoders of varying depths.}
    \scalebox{0.8}{\newdepth}
\end{table}

\subsection{Ablations around optimal configuration}
In the main paper, we sequentially address the challenges encountered by Latent MIM models, proposing a series of solutions that progressively build upon each other. However, it is possible that the effectiveness of early solutions was due to the shortcomings of the initial latent MIM implementation. To further dissect the impact of each component, we conduct ablation studies around the optimal configuration. The downstream performance of these models is detailed in \cref{tab:ablations}. As evidenced by the results, the proposed solutions effectively tackle the challenges, even when compared to robust latent MIM models that incorporate all but one of the proposed solutions.



\begin{table}[t!]
    \centering
    \caption{\textbf{Ablation studies around optimal configuration}. We report top-1 nearest neighbor (NN), linear probing (Lin) and fine-tuning (Ft) accuracy (\%) on IN100 dataset. The default configuration is marked in \colorbox{gray!10}{gray}. \label{tab:ablations}}
    \begin{subtable}{0.25\textwidth}
        \centering
        \caption{Patch Gap\label{tab:gap}}
        
        \scalebox{0.8}{\newgap}
    \end{subtable}\quad
    \begin{subtable}{0.35\textwidth}
        \centering
        \caption{Mask Ratio\label{tab:maskratio}}
        
        \scalebox{0.8}{\newmaskratio}
    \end{subtable}\quad
    \begin{subtable}{0.32\textwidth}
        \centering
        \caption{Similarity Constrain\label{tab:avgsimconstrain}}
        
        \scalebox{0.8}{\newavgsimconstrain}
    \end{subtable}
    \\
    \begin{subtable}{0.45\textwidth}
        \centering
        \caption{Decoder Design\label{tab:modeldesign}}
        
        \scalebox{0.8}{\newmodeldesign}
    \end{subtable}
    \qquad
    \begin{subtable}{0.35\textwidth}
        \centering
        \caption{Projector\label{tab:projector}}
        
        \scalebox{0.8}{\newprojector}
    \end{subtable}
    \\
    \begin{subtable}{0.4\textwidth}
        \centering
        \caption{Loss\label{tab:loss}}
        
        \scalebox{0.8}{\newloss}
    \end{subtable}
\end{table}
\section{ImageNet-1K Classification: Beyond low-level and latent MIM.}

\begin{table}[t!]
    \caption{\label{tab:in1k-cls-full}\textbf{Comparison with prior work on ImageNet-1k}.}
    \centering
    \resizebox{0.6\linewidth}{!}{\infull}
\end{table}

The main paper provides comparisons to prior low-level and latent MIM frameworks on ImageNet-1k, focusing on representation quality. \cref{tab:in1k-cls-full} provides a broader view of the downstream performance on ImageNet-1k, including comparisons to other methods beyond low-level and latent MIM, while also assessing finetuning performance. 
As can be seen, when evaluating on frozen representations (\ie, nearest neighbor classification and linear probing), the proposed latent MIM framework outperforms prior low-level and latent MIM methods, as well as MIM models using frozen pre-trained targets models. 
As for fine-tuning, although not state-of-the-art, latent MIM still achieves competitive results, showing that the proposed model can also be used to provide a strong model initialization. Since we focus our efforts on representation quality and the challenges of the latent MIM framework, we believe that a more in-depth exploration of the fine-tuning recipe could further improve the performance of our model.

The bottom two panels of~\cref{tab:in1k-cls-full} broaden the catalog of standalone MIM to (1) global joint embedding frameworks (or contrastive learning) and (2) hybrid variants that pair latent MIM with additional self-supervised learning objectives. The top-performing methods iBoT and SIM pair latent MIM with a global contrastive learning objective, thus demonstrating the benefits of combining complementary sources of supervision.
Similar to these works, our framework can also be easily combined with other self-supervised learning objectives to further improve performance. 
However, we believe that addressing latent MIM as an independent problem, without masking its shortcomings with supplementary objectives, should expedite its development, ultimately leading to more effective hybrid methods. For these reasons, while promising, we left the exploration of hybrid approaches for future work.
\section{Visualizations of Latent MIM local features}

\begin{figure}[t!]
    \centering
    \includegraphics[width=\textwidth]{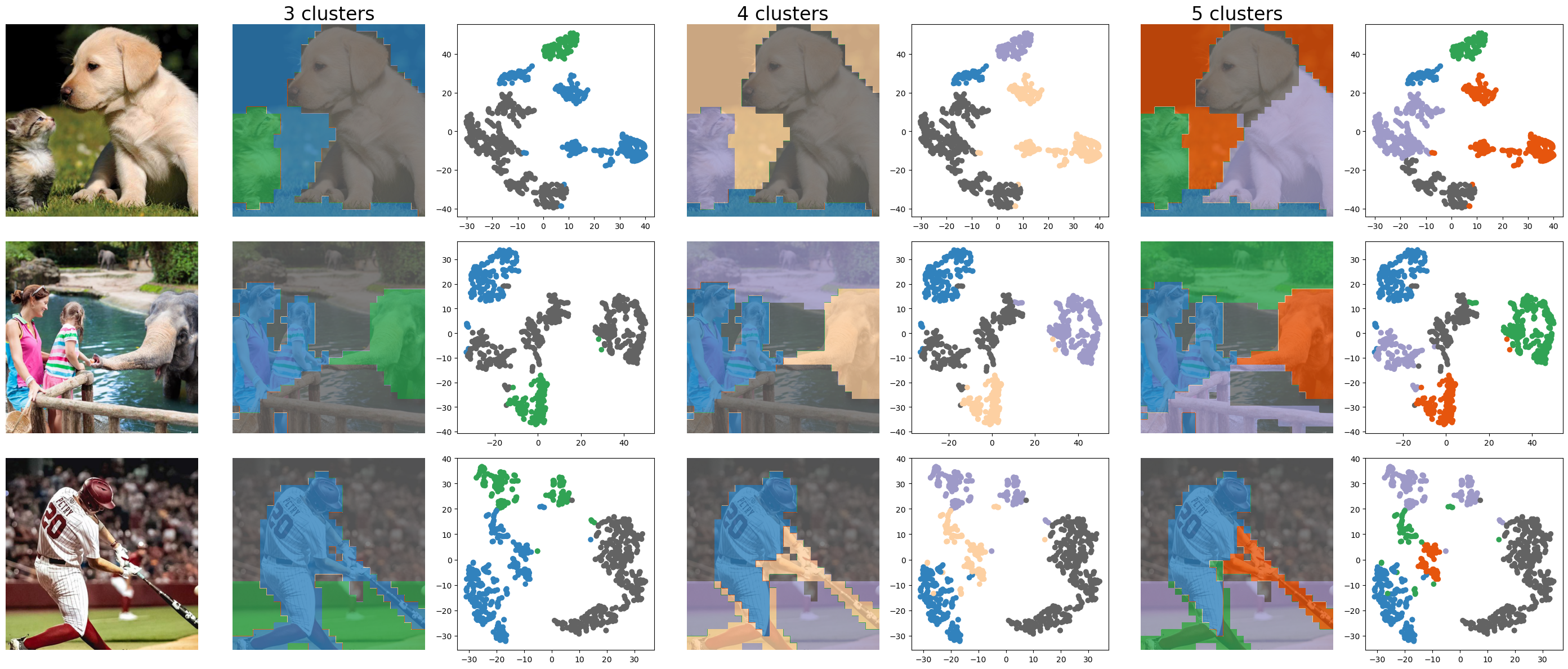}
    \caption{\label{fig:cluster-embeddings}Hierarchical clustering of local features within a single image and t-SNE embeddings of the corresponding latent features.}
\end{figure}

We showcase the localizability and semantic richness of Latent MIM features by visualizing the learned local representations within an image. All visualizations are produced using the ViT-B/16 model trained on ImageNet-1K. 

\begin{figure}[t!]
    \centering    
     \begin{subfigure}[b]{\linewidth}
        \includegraphics[width=\linewidth]{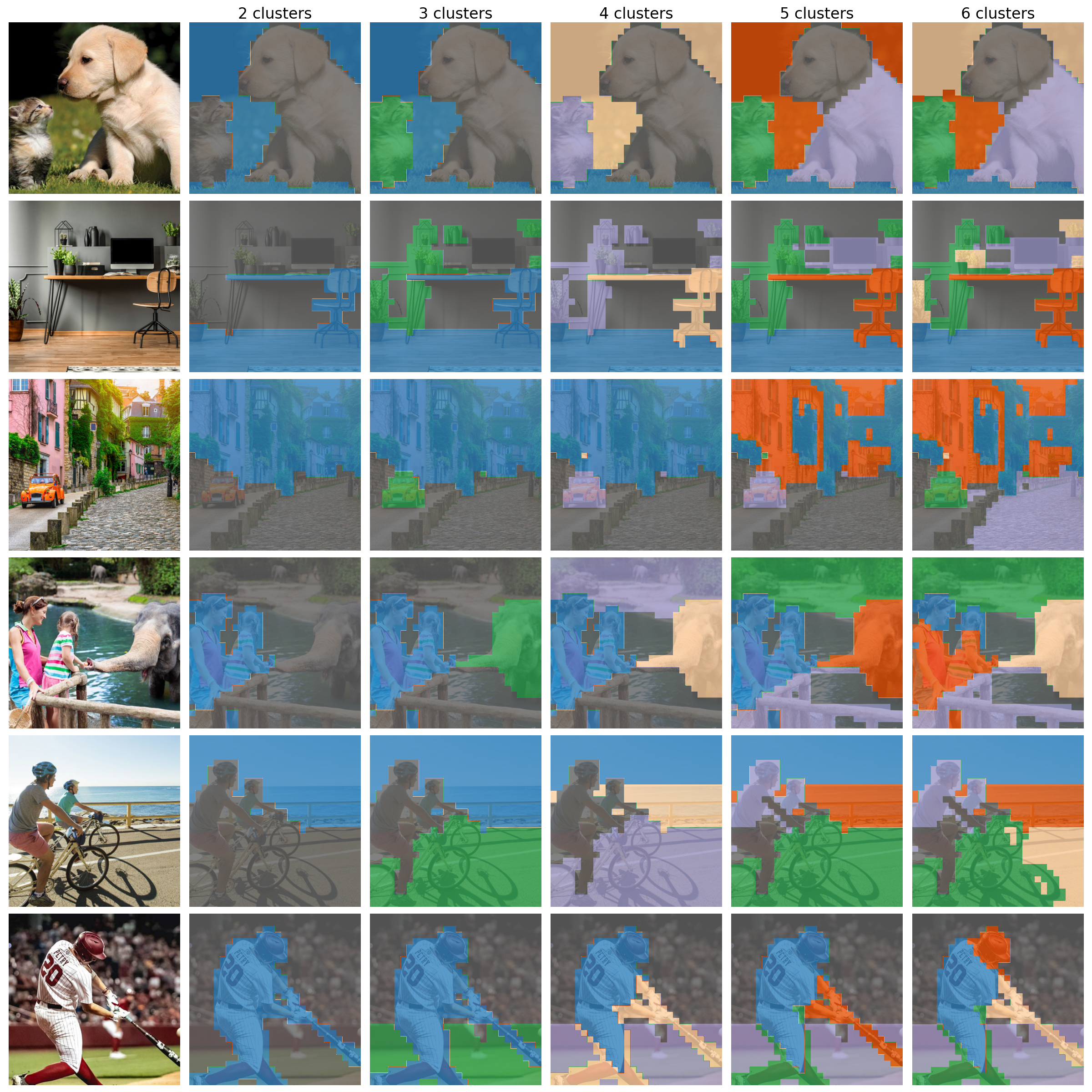}
        \caption{Ours}
    \end{subfigure}\\
     \begin{subfigure}[b]{0.47\linewidth}
        \includegraphics[width=\linewidth]{figs/data2vec-hierclt.png}
        \caption{data2vec}
    \end{subfigure}\hfill
    \begin{subfigure}[b]{0.47\linewidth}
        \includegraphics[width=\linewidth]{figs/mae-hierclt.png}
        \caption{MAE}
    \end{subfigure}
    \caption{\textbf{Unsupervised segmentation maps} by hierarchical clustering of patch representations within each image.}
    \label{fig:hier-clt}
\end{figure}

\cref{fig:hier-clt} shows unsupervised segmentation maps generated by hierarchical clustering of patch-level representations, comparing to MAE and data2vec. The results demonstrate our model's superior ability to semantically cluster pixels, compared to both lower-level MIM approaches, such as MAE, and earlier latent MIM methods like data2vec. This is consistent across various scenes. These qualitative examples also confirm our hypothesis: low-level MIM features are inadequate for grasping high-level semantics. For example, in the case of the baseball player, MAE shows segmentation maps where pixels seem to be clustered based on color, grouping the white shirt with the white signs in the background. In contrast, our model can capture the semantic boundaries of the various components of the scene.

To further understand how the latent MIM learned representations are clustered, we visualize the t-SNE embeddings of the local features within an image in \cref{fig:cluster-embeddings}. Each point in the t-SNE plot represents a patch in the image, color-coded by the hierarchical clustering labels shown in the segmentation maps next to it. As can be seen, the latent representations form semantically meaningful clusters and sub-clusters associated with objects and object parts, respectively. For example, notice, how the baseball bat, the player's body and legs are clustered separately within the larger cluster of the player.

\begin{figure}[t!]
    \centering
    \caption{\label{fig:pixel-semantic-neighbors}Semantic neighborhood of image patches. Within each image, we select 3 random pixels (highlighted in green) and visualize the regions of the image with similar representations, as measured by cosine similarity. The similarity threshold is decreased from the left-most column to the right-most column.}
    \begin{subfigure}{0.8\textwidth}
        \centering
        \includegraphics[width=\linewidth]{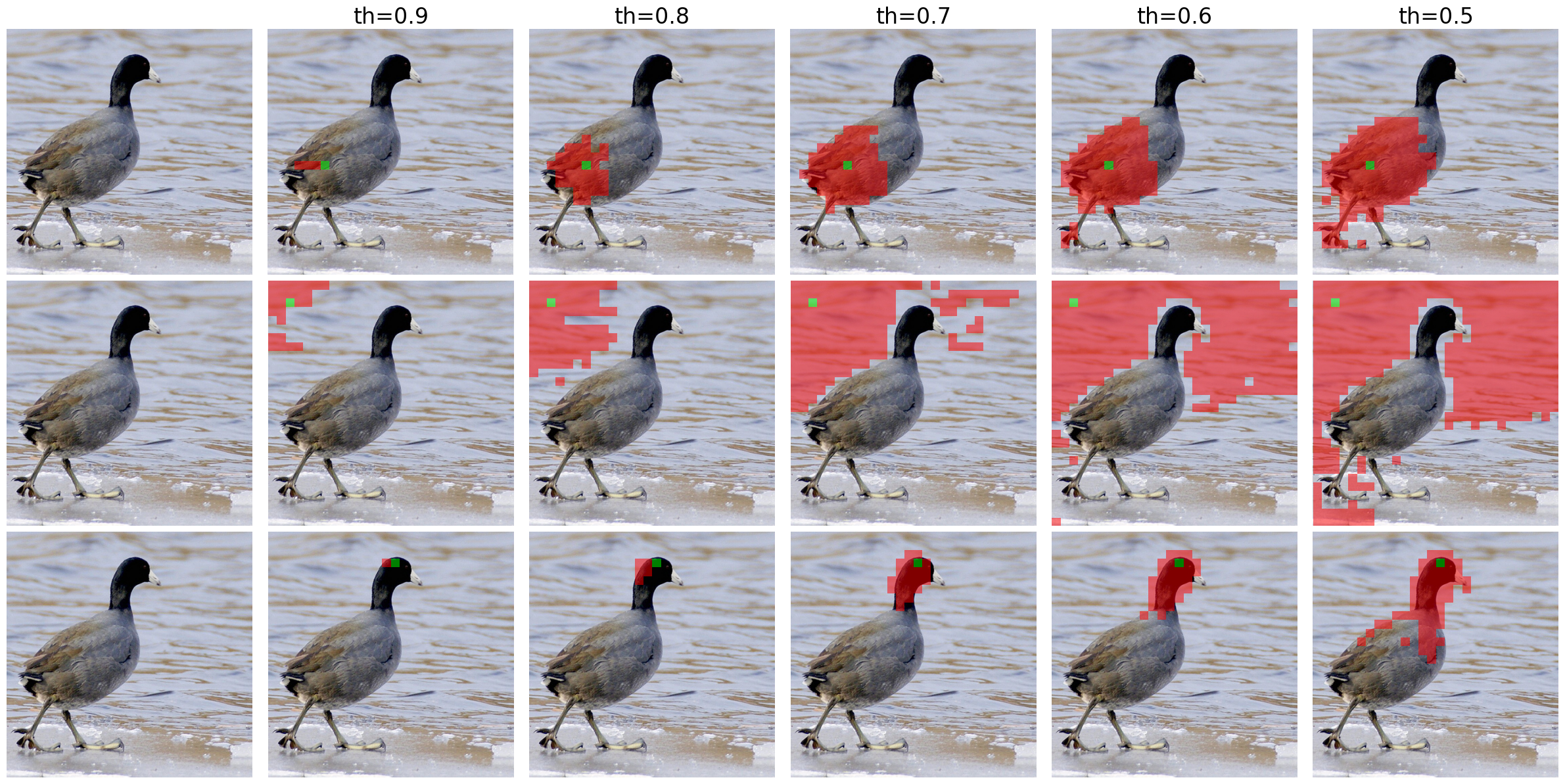}
        \caption{\label{fig:neigh1}}
    \end{subfigure}
    \begin{subfigure}{0.8\textwidth}
        \centering
        \includegraphics[width=\linewidth]{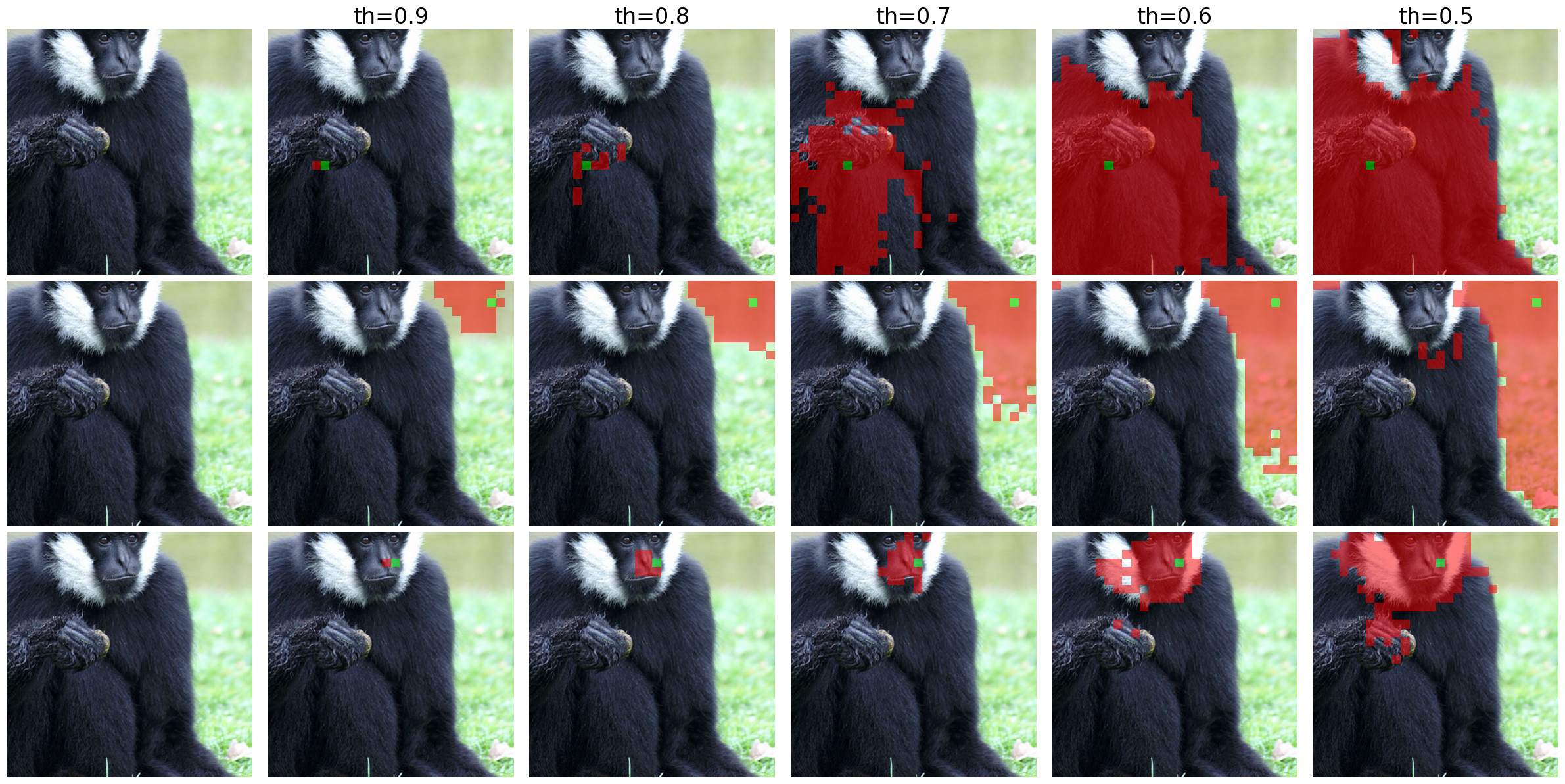}
        \caption{\label{fig:neigh1}}
    \end{subfigure}
    \begin{subfigure}{0.8\textwidth}
        \centering
        \includegraphics[width=\linewidth]{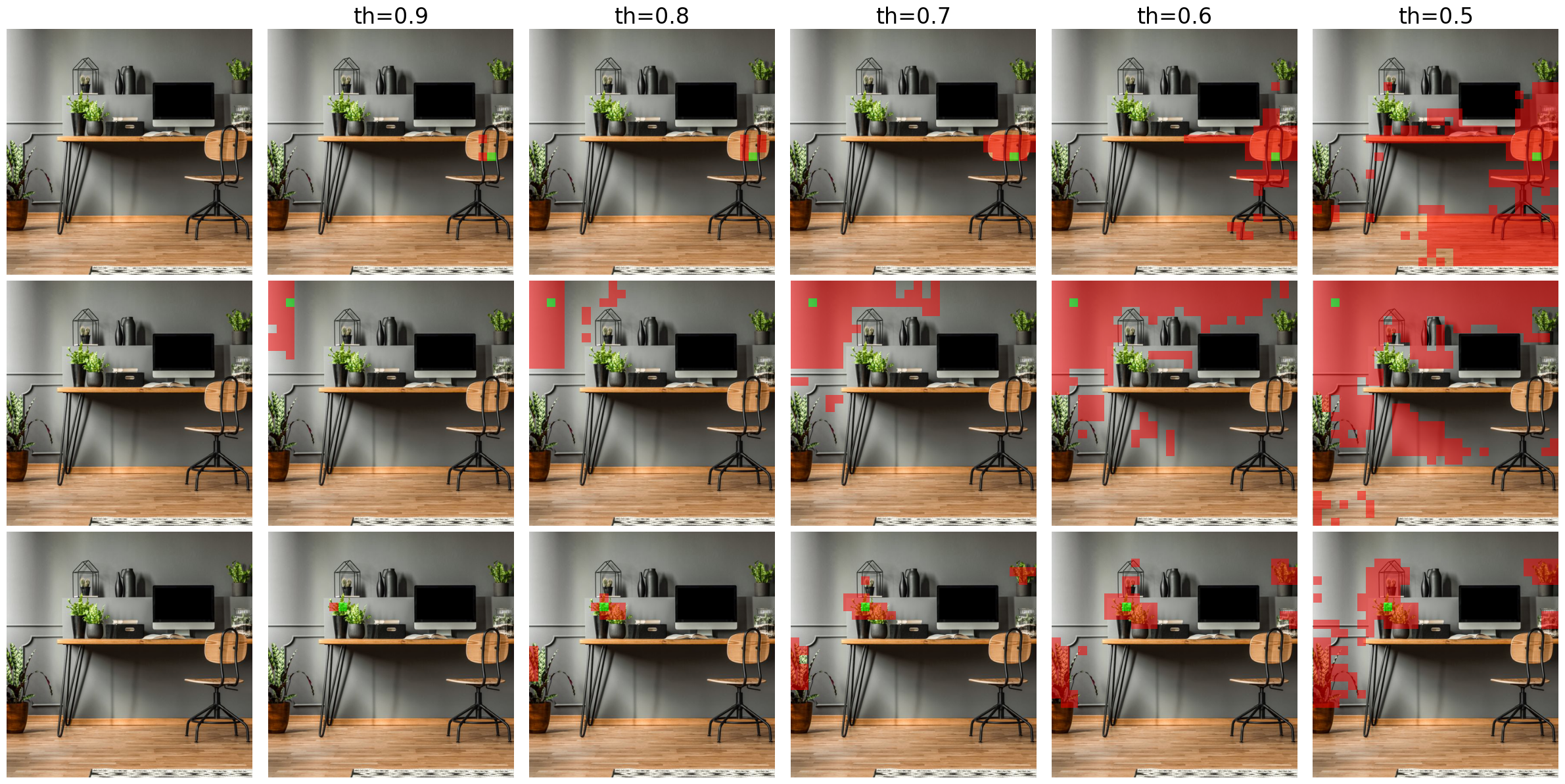}
        \caption{\label{fig:neigh1}}
    \end{subfigure}
\end{figure}

Finally, we visualize the semantic neighborhood of patches in a variety of images in \cref{fig:pixel-semantic-neighbors}. Within each image, we visualize regions with similar representations to a given pixel. We observe that, as the similarity threshold decreases, the semantic neighborhood becomes more inclusive, capturing increasingly larger regions of the image, while still preserving the semantic coherence of the neighborhood. This indicates that the learned representations encode both semantics and spatial information about the image.

\end{document}